\documentclass[10pt,twocolumn,letterpaper]{article}

\usepackage[pagenumbers]{cvpr} 

\usepackage{booktabs,multirow,tabularx,xcolor,colortbl}

\usepackage{cuted}
\usepackage{caption}
\usepackage[most]{tcolorbox}

\definecolor{qaBackground}{HTML}{F5F5FF}
\definecolor{qaBorder}{HTML}{AAAAC8}

\newtcolorbox{qabox}{
    enhanced,
    colback=qaBackground,
    colframe=qaBorder,
    boxrule=0.4pt,
    sharp corners,
    boxsep=0pt,
    left=6pt,
    right=6pt,
    top=4pt,
    bottom=4pt,
    before skip=6pt,
    after skip=6pt
}

\definecolor{groupbg}{RGB}{242,247,251}
\definecolor{oursbg}{RGB}{222,240,232}
\definecolor{gainred}{RGB}{190,45,45}
\definecolor{dropblue}{RGB}{45,90,170}

\usepackage{booktabs,multirow,tabularx,xcolor,colortbl,makecell}
\usepackage{titletoc}
\definecolor{gainred}{RGB}{190,45,45}
\definecolor{dropblue}{RGB}{45,90,170}
\definecolor{refblue}{RGB}{45,90,170}

\usepackage[normalem]{ulem}

\newcommand{\bestref}[1]{\uuline{#1}}

\newcommand{\scorediff}[1]{%
{\small%
\ifdim#1pt>0pt%
\textcolor{gainred}{$#1$}%
\else\ifdim#1pt<0pt%
\textcolor{dropblue}{$#1$}%
\else%
$#1$%
\fi\fi}}

\definecolor{cvprblue}{rgb}{0.21,0.49,0.74}
\usepackage[pagebackref,breaklinks,colorlinks,allcolors=cvprblue]{hyperref}

\def\confName{CVPR}
\def\confYear{2027}

\title{Beyond Attention Maps: Understanding and Improving Medical MLLMs via Visual Values}

\usepackage{etoolbox}

\makeatletter
\patchcmd{\@maketitle}
  {\vspace*{12pt}}
  {\vspace*{4pt}}
  {}
  {\PackageWarning{title-spacing}{Author spacing patch failed}}

\patchcmd{\@maketitle}
  {\vspace*{24pt}}
  {\vspace*{12pt}}
  {}
  {\PackageWarning{title-spacing}{Title spacing patch failed}}
\makeatother

\author{Yichen Zhao$^{1,2,3}$\thanks{Equal contribution.} \qquad
Fenghe Tang$^{1,2,3}$\footnotemark[1] \qquad 
Wenxin Ma$^{4}$ \\
Rongsheng Wang$^{1,2,3}$ \qquad 
Zihao Wan$^{1,2,3}$ \qquad 
Shaohua Kevin Zhou$^{1,2,3}$\thanks{Corresponding author.}
\\
$^1$ School of Biomedical Engineering, Division of Life Sciences and Medicine, USTC\\
$^2$ MIRACLE Lab, Suzhou Institute for Advance Research, USTC \\
$^3$ Jiangsu Key Laboratory of Multimodal Digital Twin Technology, Suzhou, Jiangsu, China 215123\\
$^4$  The University of Texas at Austin \\
{\tt\small qiuzyc@mail.ustc.edu.cn fhtan9@mail.ustc.edu.cn skevinzhou@ustc.edu.cn}
}

\begin{document}
\maketitle
\begin{abstract}

Current MLLM methods largely focus on whether models attend to the right visual regions, implicitly treating spatial attention map as the primary key of effective visual perception. To examine this assumption,  we introduce a medical VQA dataset with paired non-contrast (NC)  and contrast-enhanced (CE) images that share corresponding anatomy while providing different cues. We show that this view is incomplete: across paired NC and CE medical images, MLLMs often exhibit similar attention patterns despite producing contrasting answers, while controlled bidirectional interventions induce substantially larger prediction changes on visual values than attention weights. These results identify the visual content retrieved through values as a critical factor in improving perception beyond spatial attention alone. Guided by this finding, we propose \textbf{VIVID} (\underline{VI}sual \underline{V}alue \underline{I}ntervention \underline{D}istillation), which uses enhanced visual values to identify and refine teacher corrections aligned with useful visual evidence to improve on-policy distillation from CE to NC images. Across six internal and external test sets, VIVID consistently outperforms existing on-policy distillation methods across different model scales and architectures, requires only NC images at inference, and maintains strong zero-shot performance on general-domain VQA benchmarks. The project page is available at \url{https://qiuzyc.github.io/VIVID}.

\end{abstract}    
\section{Introduction}
\label{sec:intro}

\begin{figure}[t]
    \centering
    \includegraphics[width=\linewidth]{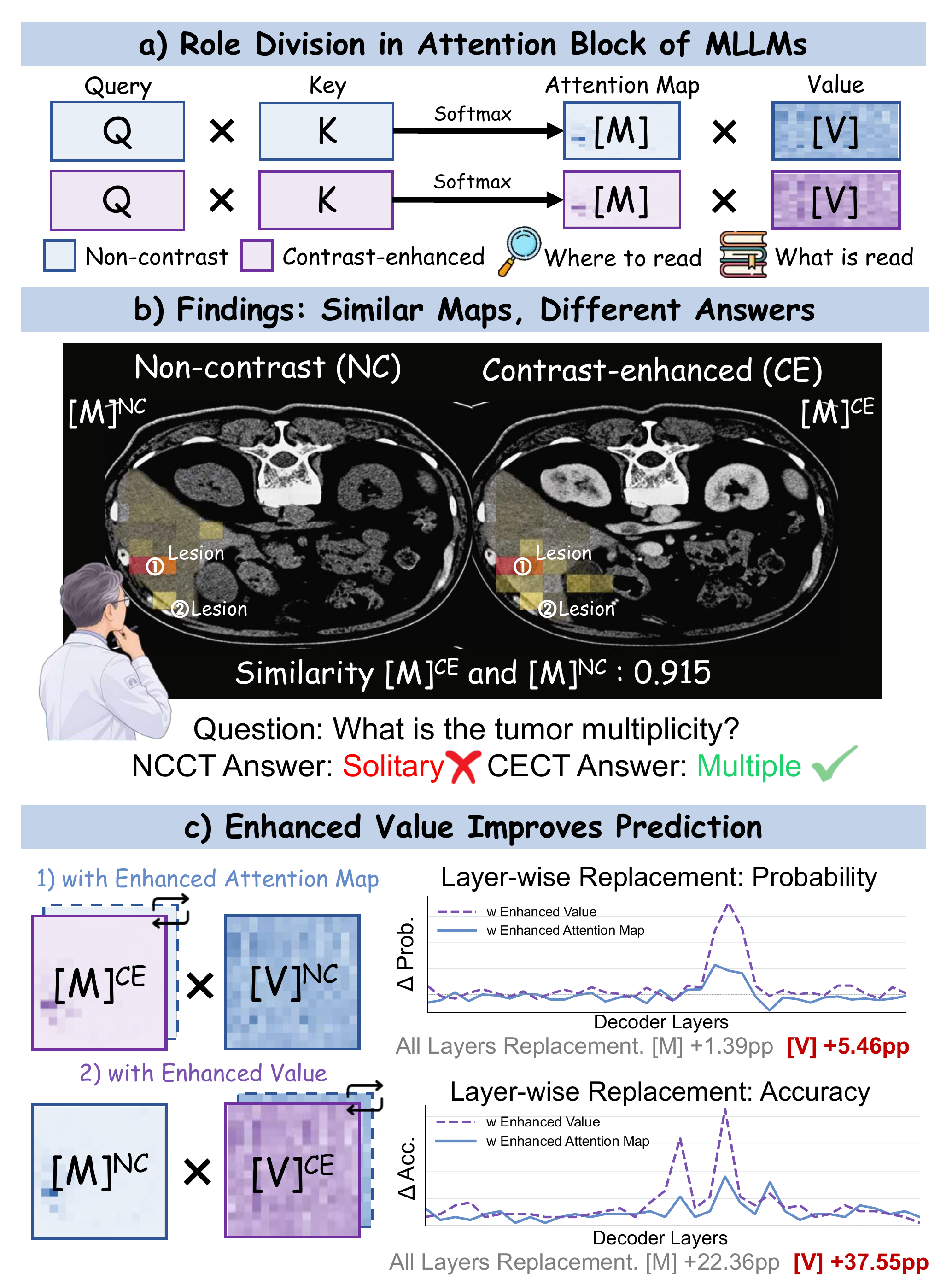}
    \caption{\textbf{Visual values improve predictions more than attention maps.} (a) \texttt{[M]} determines where to read, while \texttt{[V]} provides what is read. (b) Paired NC and CE images yield similar attention maps but different answers. (c) CE-to-NC interventions compare transferring \texttt{[M]} versus \texttt{[V]}, with larger gains from \texttt{[V]} under both single-layer and all-layer replacement.}
    \vspace{-4mm}
    \label{fig:teaser}
\end{figure}

Multimodal large language models (MLLMs) can produce inaccurate responses when they under-attend to visual inputs or focus on irrelevant image regions~\cite{liu2024paying,gong2024damro}. To improve visual understanding, prior work has strengthened attention to relevant image tokens, suppressed distracting visual tokens, or aligned attention with relevant regions~\cite{liu2024paying,gong2024damro,chang2025focus}. These methods share a common focus: they shape the attention weights, denoted as \texttt{[M]}, over visual tokens which can be visualized as a spatial attention map.
However, attention weights are only one half of the mechanism. As shown in Figure~\ref{fig:teaser}, panel A, while \texttt{[M]} determines \emph{where} to attend, the visual value representations, denoted as \texttt{[V]}, determine \emph{what} information is retrieved: the attention readout is the \texttt{[M]}-weighted aggregation of \texttt{[V]}~\cite{kobayashi2020attention}. Existing guidance approaches regulate \texttt{[M]} while leaving \texttt{[V]} unexamined. Given that refining \texttt{[M]} yields only limited gains in grounding accuracy, this raises a question: \textbf{\textit{Is visual grounding limited by where the model looks, or by what it retrieves?}} 

Answering this question requires disentangling spatial attention patterns from the visual information available at the attended locations. Paired non-contrast and contrast-enhanced medical images provide a natural testbed for this purpose: they offer anatomically corresponding views while differing in the visual cues revealed by contrast enhancement~\cite{shaish2023diagnostic,gong2018deep}. Leveraging this property, we construct a paired medical VQA diagnostic dataset spanning three anatomical sites and two imaging modalities, where registered image pairs allow the same patient and question to be examined under different visual evidence conditions. 
Across these paired views, we find that the same model can answer incorrectly from a non-contrast image yet correctly from its enhanced counterpart, despite exhibiting similar spatial attention distributions, as shown in Figure~\ref{fig:teaser}, panel B. This discrepancy suggests that \emph{where} the model attends alone cannot fully explain its grounding behavior, motivating us to examine \emph{what} visual information is retrieved at the attended locations and to leverage the enhanced view as a source of privileged visual supervision.

Leveraging the spatial correspondence between paired scans, we disentangle \texttt{[M]} and \texttt{[V]} through bidirectional substitution between NC and CE views, with the
CE-to-NC direction shown in Figure~\ref{fig:teaser}, panel C. 

Across multiple benchmarks, replacing \texttt{[V]} produces larger changes in model predictions than \texttt{[M]}, with directionally consistent effects: enhanced values increase support for the correct answer in the non-contrast branch, while non-contrast values reduce it in the enhanced branch. These controlled interventions indicate that the performance gap between the two views is driven primarily by differences in the visual information carried by \texttt{[V]}, rather than by differences in spatical allocation encoded by \texttt{[M]}. 
In other words, \textbf{\emph{what} the model retrieves can matter more than \emph{where} it attends.}

Motivated by these findings, we investigate how enhanced visual values can guide knowledge transfer from a contrast-enhanced teacher to a non-contrast student. While an enhanced-image teacher can provide richer supervision, privileged information does not necessarily translate into effective student supervision, as shown in prior work on thinking models~\cite{kaur2026rethinking}. In our setting, this motivates examining whether the teacher's corrections align with the student's response to answer-relevant visual evidence.
To allow reasonable \texttt{[V]}-enhanced distillation, we build on on-policy self-distillation (OPSD)~\cite{zhao2026self} and propose \textbf{VIVID} (\underline{VI}sual \underline{V}alue \underline{I}ntervention \underline{D}istillation). We introduce an auxiliary branch that shares the student's non-contrast input and generated answer prefix, while incorporating paired enhanced visual values \texttt{[V]} into the visual readout at attended image locations. The resulting prediction shift captures the student's response to enhanced visual evidence and serves as a reference direction for the teacher's correction. We project the enhanced-image teacher's correction onto this direction and retain the positively aligned component as the principal distillation target. At inference, only the non-contrast student is required. Experiments on diverse VQA datasets show that VIVID consistently outperforms off-the-shelf MLLMs and on-policy distillation baselines.

Our contributions are summarized as follows:

\begin{enumerate}
    \item \textit{Revealing what models retrieve beyond where they attend.}
    We establish the first paired medical VQA setting with registered NC and CE scans across three anatomical sites and two imaging modalities. Controlled interventions show that visual values \texttt{[V]} have a substantially greater influence on predictions than attention weights \texttt{[M]}., revealing retrieved visual content as a critical factor beyond spatial attention patterns.

    \item \textit{Visual value-guided distillation for non-contrast inference.}
    We propose \textbf{VIVID}, which uses enhanced visual values to better identify relevant teacher corrections and guide knowledge transfer from contrast-enhanced to non-contrast images, while requiring only non-contrast images at inference.

    \item \textit{Effective and generalizable medical VQA.}
    Extensive experiments across anatomical sites, imaging modalities, and MLLMs architectures demonstrate consistent improvements over existing off-the-shelf MLLMs and on-policy distillation approaches, together with strong zero-shot performance on general-domain VQA benchmarks.
\end{enumerate}
\section{Related Work}
\label{sec:related_work}

\subsection{Attention and Visual Information in MLLMs}

A major line of work improves visual understanding in MLLMs by directing attention toward relevant image regions. PAI amplifies image-token attention with contrastive logit correction~\cite{liu2024paying}, while DAMRO suppresses high-attention background outliers~\cite{gong2024damro}. In medical VQA, Localize-before-Answer localizes pathological regions to guide answer generation~\cite{nguyen2025localizing}. These approaches emphasize where models attend, leaving the answer-relevant content retrieved at those locations less directly examined. Earlier Transformer analyses also demonstrate that attention weights alone do not characterize token contributions, highlighting the role of transformed value-vector norms~\cite{kobayashi2020attention}. We examine this distinction through paired NC--CE medical images with corresponding anatomy but different diagnostic cues. Despite similar spatial attention patterns, models can produce different answers across views; controlled interventions further reveal that visual values transfer the CE answer advantage more effectively than attention weights. This finding motivates using enhanced visual values to guide learning from non-contrast images.

\subsection{Distillation with Privileged Visual Information}

Contrast administration adds cost and risks adverse reactions~\cite{kodzwa2019acr}, motivating knowledge transfer from enhanced to non-contrast images. Prior work focuses on segmentation: HAD-Net distills full-sequence MRI knowledge into a student without post-contrast inputs~\cite{vadacchino2021had}, while EGTA-KD transfers enhanced-image knowledge to non-contrast tumor segmentation~\cite{zhao2025evidence}. Such transfer remains less explored in generative medical VQA. On-policy distillation aligns teacher and student predictions along student-generated sequences~\cite{agarwal2024policy}. Vision-OPD transfers crop-conditioned perception to a full-image student~\cite{yuan2026vision}. VA-OPD uses original/degraded-image comparisons to reweight rollouts and balance token groups~\cite{liu2026visual}, while Med-OPD uses answer-aware hints to emphasize medically relevant tokens and trajectories~\cite{qian2026med}. While these approaches improve distillation by adjusting the importance of supervision based on visual dependence or task-relevant evidence, they do not explicitly assess whether teacher corrections align with the student's response to privileged visual information. In contrast, VIVID intervenes on visual values using paired CE images to obtain an evidence-induced prediction direction, then retains the positively aligned component of the teacher's correction to construct refined distillation targets.
\section{Diagnostic Analysis of Visual Understanding}

To examine whether visual understanding is limited by where the model looks or by what information it extracts, we first introduce a paired non-contrast and contrast-enhanced medical VQA dataset (Sec.~\ref{sec:benchmark}). We then analyze attention patterns and perform controlled attention map and value interventions to assess their respective roles in answer prediction (Secs.~\ref{sec:value_analysis} and~\ref{sec:value_interventions}).


\subsection{Paired Diagnostic Dataset Construction}
\label{sec:benchmark}

We construct a paired medical VQA dataset covering liver CT, brain MRI, and breast MRI.

\noindent\textbf{Data Sources.} For liver CT, we use MCT-LTDiag~\cite{wu2025multi} for training and internal evaluation and TriALS~\cite{elbatel2026trials} for external evaluation. For brain MRI, we train and internally evaluate on cohorts including UPENN-GBM~\cite{bakas2022university} and UCSF-PDGM~\cite{calabrese2022university} and other four datasets, with TCGA-GBM~\cite{scarpace2016tcgagbm} and TCGA-LGG~\cite{pedano2016tcgalgg} held out for external evaluation. For breast MRI, we use Duke~\cite{saha2018machine}, I-SPY1~\cite{newitt2016ispy1}, and I-SPY2~\cite{li2022ispy2} for training and internal evaluation and NACT~\cite{newitt2016nact} for external evaluation. All splits are patient-disjoint. 
Detailed data sources, statistics, construction procedures, and ethics considerations are provided in Appendices~\ref{app:data_sources}, \ref{app:image_preparation}.

\noindent\textbf{Task design.}
We define three clinically motivated tasks: lesion identification, lesion characterization, and multi-region localization, covering both global and local visual discrimination. We construct QA and supporting evidence at the NC--CE pair level: questions and reference answers are derived from segmentation annotations, with the same question, options, and answer assigned to both views of each pair. Concise supporting evidence is generated jointly from the paired views, yielding responses in the \texttt{Evidence: ... Answer: ...} format. Samples undergo clinical quality review (see below). Detailed construction process is provided in Appendices \ref{app:slice_question_construction} and \ref{app:evidence_generation}.

\noindent\textbf{Clinial quality assurance.} We conduct radiologist-led review before and after full-scale data generation. During the pilot stage, a radiologist evaluates generated evidence for spatial consistency and factual accuracy with answer support using two five-point Likert scales. The findings guide revisions to the generation instructions to reduce unsupported inferences and preserve region-specific descriptions. After full-scale generation, the same radiologist reviews a 20\% sample using the same criteria, obtaining mean scores of 4.6/5.0 and 4.5/5.0, respectively. Detailed procedures and dataset cases are provided in Appendices \ref{app:clinical_review} and \ref{app:dataset_cases}.

\begin{figure*}[t]
\centering
\includegraphics[width=\textwidth]{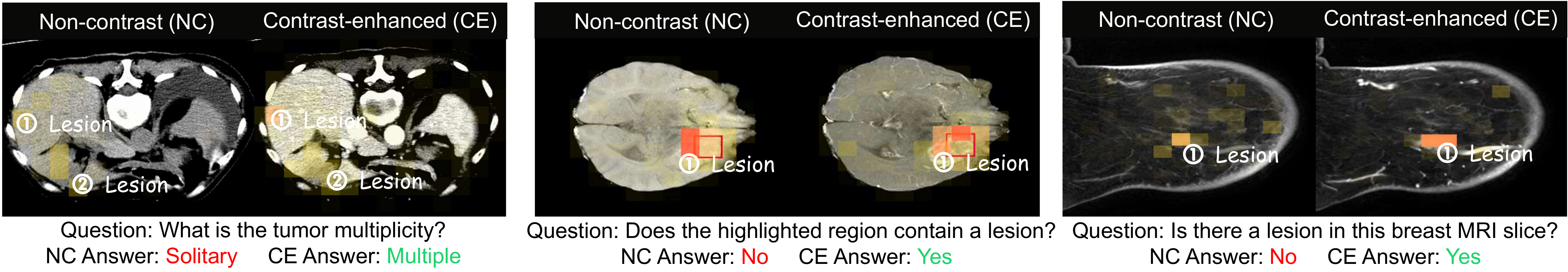}
\caption{\textbf{Similar attention map, different answers.} Paired NC and CE scans from liver, brain, and breast cases. In each example, the model exhibits similar attention patterns but answers incorrectly from the NC scan and correctly from the CE scan.}
\label{fig:attention_examples}
\end{figure*}

\subsection{What Matters to Answering Quality}
\label{sec:value_analysis}

Given frozen models, our analysis focuses on attention to image tokens within the MLLM's decoder during answer generation. At each prediction position, a textual query \texttt{[Q]} attends to image-token keys \texttt{[K]}, producing spatial attention weights \texttt{[M]}. These weights aggregate the corresponding visual values \texttt{[V]} to form the visual contribution to the attention readout before the output projection.

\begin{figure}[t]
\centering
\includegraphics[width=\columnwidth]{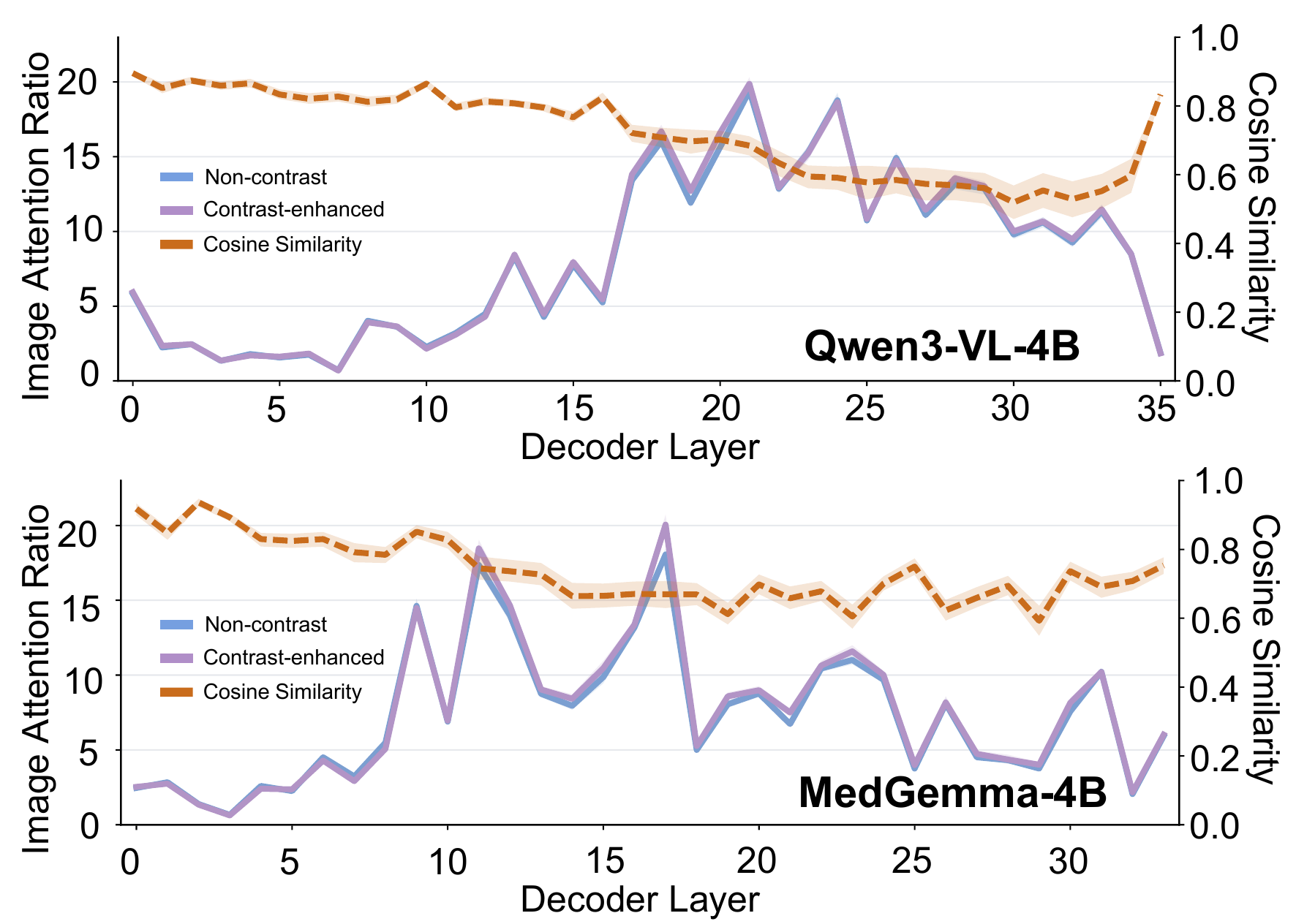}
\caption{\textbf{Layer-wise attention patterns on liver CT.} Blue and purple curves show the fraction of attention weight allocated to image tokens for NC and CE inputs (left axis). The orange curve shows cosine similarity between their visual attention maps (right axis). Shaded bands indicate 95\% confidence intervals.}
\label{fig:attention_curves}
\end{figure}

\begin{qabox}
\textbf{Finding 1: Spatial attention alone does not explain contrasting predictions.}
\end{qabox}

We analyze attention maps in MLLMs on paired samples where the NC branch answers incorrectly while the CE branch answers correctly, with each branch generating independently. Despite the difference in prediction correctness, their spatial attention patterns remain remarkably similar. Qualitatively, Fig.~\ref{fig:attention_examples} shows closely matched attention maps across liver CT, brain MRI, and breast MRI examples. Quantitatively, on liver CT, the fraction of attention assigned to image tokens follows nearly identical layer-wise trends between NC and CE inputs in both Qwen3-VL and MedGemma. Foreground attention maps also exhibit high cosine similarity after excluding the black image background (Fig.~\ref{fig:attention_curves}). We observe consistent patterns on the brain and breast MRI benchmarks. These results suggest that differences in \emph{where} the model attends alone do not explain the contrasting predictions between paired NC and CE views. Detailed protocols and results for brain and breast MRI are provided in Appendix \ref{app:attention_similarity}.

The similarity of NC and CE attention maps raises a natural question: \textit{if \texttt{[M]} does not explain their contrasting predictions, does the difference lie in the visual information carried by \texttt{[V]}? }

\subsection{Visual Values Shape Answer Predictions}
\label{sec:value_interventions}
We answer above question through controlled, bidirectional exchanges of \texttt{[M]} and \texttt{[V]} between paired views.

\noindent\textbf{Experimental setting.} 
To investigate the answer gap between paired NC and CE views, we focus on questions that frozen SFT models answer incorrectly from NC images but correctly from CE images. This subset allows us to test which component transfers the CE branch's answer advantage. We exchange either \texttt{[M]} or \texttt{[V]} between paired branches while keeping the other component unchanged, and measure changes in correct-answer probability $P_{\mathrm{GT}}$ and answer accuracy Acc. We report all-layer results on liver CT here. Analyses of other subsets defined by answer correctness, single-layer interventions, and results for brain and breast MRI are provided in Appendix~\ref{app:paired_interventions}.

\begin{qabox}
\noindent\textbf{Finding 2: Predictions are substantially more sensitive to \texttt{[V]} than to \texttt{[M]}.}
\end{qabox}

Table~\ref{tab:paired_exchange} shows a clear asymmetry between \texttt{[M]} and \texttt{[V]}. Transferring CE values to the NC branch produces substantially larger gains than transferring CE attention weights, while transferring NC values to the CE branch causes substantially larger degradation. This pattern holds for both Qwen3-VL and MedGemma and across both evaluation protocols. The bidirectional effects indicate that \texttt{[V]} carries much of the answer-relevant difference between paired views that is not captured by their spatial attention patterns.

We further conduct a stress test to prove that the advantage of CE values does not merely depend on being combined with a well-formed NC attention map. After transferring CE values to the NC branch, we progressively perturb its image-attention weights with Gaussian noise while preserving nonnegativity and total image-attention mass. Even under attention weight perturbations, CE values retain gains over the original NC prediction and remain competitive with, or stronger than, transferring CE attention weight alone (Table~\ref{tab:noisy_attention}). Although performance decreases as attention becomes noisier, this stress test shows that informative values remain beneficial even when spatial attention becomes less reliable. Thus, attention facilitates the use of visual values, but does not by itself account for their predictive advantage.

\begin{table}[t]
\centering
\caption{\textbf{Paired \texttt{[M]} and \texttt{[V]} exchanges on liver CT.} \texttt{[M]} and \texttt{[V]} denote spatial attention weights and visual values. Results are reported in percentage points. Arrows indicate donor-to-recipient transfer. Red $\Delta$ entries report \texttt{[V]} minus \texttt{[M]}.}
\label{tab:paired_exchange}
\small
\setlength{\tabcolsep}{3pt}
\renewcommand{\arraystretch}{0.95}
\setlength{\aboverulesep}{0.2ex}
\setlength{\belowrulesep}{0.25ex}
\resizebox{0.98\linewidth}{!}{%
\begin{tabular}{@{\hspace{3pt}}lccccc@{\hspace{3pt}}}
\toprule
\multirow{2}{*}{Model} & \multirow{2}{*}{Swap} & \multicolumn{2}{c}{$\Delta P_{\mathrm{GT}}$} & \multicolumn{2}{c}{$\Delta$Acc.} \\
\cmidrule(lr){3-4}\cmidrule(lr){5-6}
& & CE$\to$NC & NC$\to$CE & CE$\to$NC & NC$\to$CE \\
\midrule
\multirow{3}{*}{Qwen3-VL} & \texttt{[M]} & +1.39 & -1.61 & +22.36 & -8.02 \\
& \texttt{[V]} & \textbf{+5.46} & \textbf{-4.72} & \textbf{+37.55} & \textbf{-24.89} \\
& $\Delta$ & \textcolor{gainred}{+4.07} & \textcolor{gainred}{-3.11} & \textcolor{gainred}{+15.19} & \textcolor{gainred}{-16.87} \\
\midrule
\multirow{3}{*}{MedGemma} & \texttt{[M]} & +1.82 & -0.76 & +34.76 & -12.30 \\
& \texttt{[V]} & \textbf{+3.69} & \textbf{-4.67} & \textbf{+57.75} & \textbf{-32.09} \\
& $\Delta$ & \textcolor{gainred}{+1.87} & \textcolor{gainred}{-3.91} & \textcolor{gainred}{+22.99} & \textcolor{gainred}{-19.79} \\
\bottomrule
\end{tabular}%
}
\end{table}

\begin{table}[t]
\centering
\caption{\textbf{Value transfer under attention perturbations on liver CT.}
All interventions are applied to the NC branch.
The first two columns specify the sources of attention weights \texttt{[M]} and visual values \texttt{[V]}; noise is applied only to \texttt{[M]}.
Changes relative to the original NC predictions are reported in percentage points.}
\label{tab:noisy_attention}
\small
\setlength{\tabcolsep}{2pt}
\renewcommand{\arraystretch}{0.95}
\setlength{\aboverulesep}{0.2ex}
\setlength{\belowrulesep}{0.25ex}

\begin{tabularx}{\linewidth}{
    @{}ccc
    *{4}{>{\centering\arraybackslash}X}
    @{}
}
\toprule
\multirow{2}{*}{\texttt{[M]}}
& \multirow{2}{*}{\texttt{[V]}}
& \multirow{2}{*}{\shortstack{Noise on\\\texttt{[M]}}}
& \multicolumn{2}{c}{\textit{Qwen3-VL}}
& \multicolumn{2}{c}{\textit{MedGemma}} \\
\cmidrule(lr){4-5}\cmidrule(lr){6-7}
& & & $\Delta P_{\mathrm{GT}}$ & $\Delta$Acc.
    & $\Delta P_{\mathrm{GT}}$ & $\Delta$Acc. \\
\midrule
CE & NC & 0\%
& +1.39 & +22.36 & +1.82 & +34.76 \\
\midrule
NC & CE & 0\%
& \textbf{+5.46} & \textbf{+37.55}
& \textbf{+3.69} & \textbf{+57.75} \\
NC & CE & 25\%
& +4.66 & +34.60 & +3.64 & +54.55 \\
NC & CE & 50\%
& +3.81 & +32.07 & +3.44 & +49.20 \\
\bottomrule
\end{tabularx}
\end{table}

Finally, we test the robustness of our findings across different basem models. We separately fine-tune NC and CE Qwen3-VL-4B models from the same initialization under identical training configurations. Despite this independent training, their foreground attention maps remain similar (mean cosine similarity: 0.651), and the same intervention pattern persists: exchanging \texttt{[V]} produces consistently larger bidirectional changes in correct-option probability and accuracy than exchanging \texttt{[M]}. Detailed results are provided in the Appendix \ref{app:consistency}.

Together, these analyses show that spatial attention maps provide only a partial account of visual grounding and understanding: \texttt{[M]} determines \emph{where} visual information is aggregated, but the answer-relevant differences between paired views are expressed more strongly in \texttt{[V]}, which determines \emph{what} information is retrieved.

\section{Method}
\label{sec:method}

\begin{figure*}[t]
    \centering
    \includegraphics[width=\textwidth]{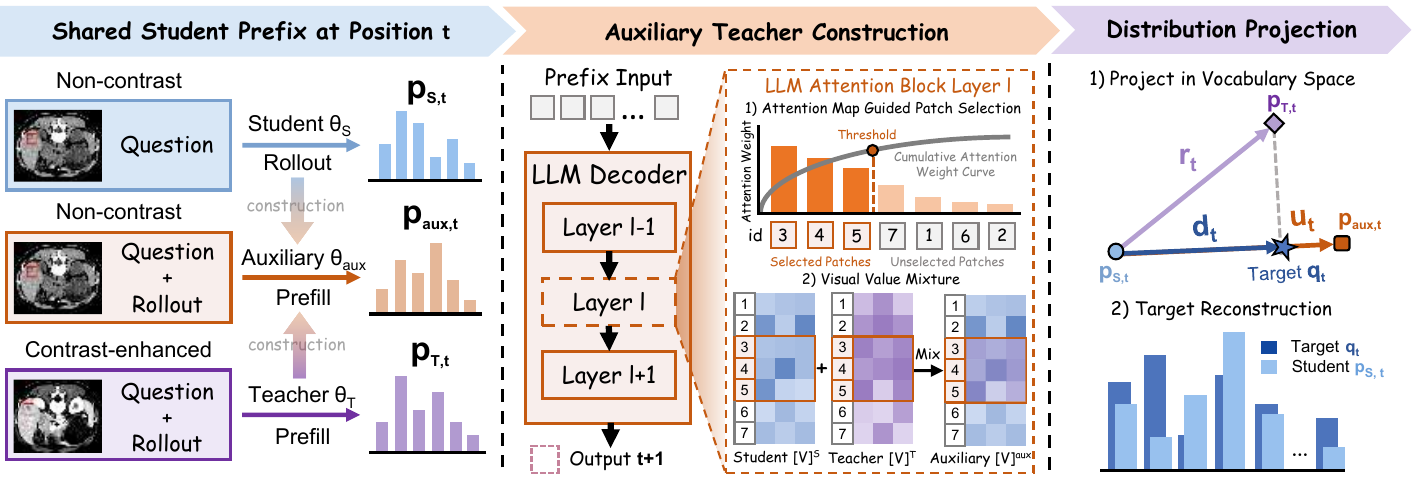}
    \caption{\textbf{Overview of VIVID.}
All branches share prefixes from the non-contrast student's rollouts. An auxiliary branch mixes paired enhanced-image values into attention-weight-selected visual positions. The induced prediction shift guides a positive projection of the teacher's correction in centered log-probability space to construct the distillation target.}
    \label{fig:method}
\end{figure*}

Building on our finding that visual values capture answer-relevant differences beyond spatial attention, we revisit how knowledge is distilled from a contrast-enhanced teacher to a non-contrast student. Existing distillation methods transfer the teacher's prediction-level corrections without explicitly accounting for whether these corrections align with the enhanced visual evidence that benefits the student~\cite{zhao2026self}. We therefore extend on-policy self-distillation (OPSD) with \textbf{Visual Value Intervention Distillation (VIVID)}, using enhanced visual values to refine targets construction.

As illustrated in Figure~\ref{fig:method}, an auxiliary branch injects paired CE visual values into the NC student along its on-policy trajectories, as described in~\cref{sec:value_mixing}. The resulting prediction shift captures the student's response to enhanced visual evidence and serves as a reference direction for refining the target. We retain the positively aligned component of the teacher's correction as the distillation signal, thereby using visual values to guide \emph{what} is distilled rather than distilling them directly, as described in~\cref{sec:target_construction}. The objective is in~\cref{sec:distillation_objective}. At inference, only the NC student is required.

\subsection{Paired Visual Value Mixing}
\label{sec:value_mixing}

At each decoder layer $\ell$, the auxiliary branch computes attention weights from its current hidden states before the intervention. At prediction position $t$, we average the weights assigned to each non-background image token across heads, rank these tokens by decreasing averaged weight, and select the smallest set whose cumulative weight reaches at least a fraction $\tau$ of the total over non-background image tokens. The selected set $\mathcal{S}_{\ell,t}$ is shared across heads. For each head $h$, we modify the attention output at position $t$ as
\begin{equation}
\widetilde{o}^{\mathrm{aux}}_{\ell,t,h} = o^{\mathrm{aux}}_{\ell,t,h} + \alpha \sum_{j \in \mathcal{S}_{\ell,t}} M^{\mathrm{aux}}_{\ell,t,h,j} \left(V^T_{\ell,h,j} - V^S_{\ell,h,j}\right),
\label{eq:paired_value_mixing}
\end{equation}
where $M^{\mathrm{aux}}_{\ell,t,h,j}$ is the attention weight assigned by prediction position $t$ to image token $j$, $V^S$ and $V^T$ are the paired NC student and CE teacher visual values, and $\alpha \in [0,1]$ controls the mixing strength. The outputs $o^{\mathrm{aux}}$ and $\widetilde{o}^{\mathrm{aux}}$ correspond to the head outputs before and after intervention, respectively. This replaces each selected value in the weighted aggregation at position $t$ with $(1-\alpha)V^S+\alpha V^T$, keeping the current attention weights unchanged. Applying this operation across all decoder layers yields $p_{\mathrm{aux},t}$ under the same student-generated prefix. Its deviation from the original student prediction $p_{S,t}$ guides target construction in the next subsection.

\subsection{Response-guided Target Construction}
\label{sec:target_construction}

We use the prediction shift induced by visual value mixing at the prediction position to guide teacher supervision. At each student-generated prefix, let $p_{S,t}$, $p_{T,t}$, and $p_{\mathrm{aux},t}$ denote the NC student, CE teacher, and auxiliary predictions. Each distribution is represented over the same $K+1$ bins: the student's top-$K$ tokens and a residual bin aggregating each distribution's probability mass over remaining tokens.

In centered log-probability space, we define the teacher correction $r_t$ and auxiliary response $u_t$ as
\begin{equation}
\begin{aligned}
r_t &= C\!\left(\log p_{T,t}-\log p_{S,t}\right),\\
u_t &= C\!\left(\log p_{\mathrm{aux},t}-\log p_{S,t}\right),
\end{aligned}
\label{eq:response_directions}
\end{equation}
where $C(v)=v-\operatorname{mean}(v)$. We compute a numerically stabilized projection of $r_t$ onto $u_t$, retaining only positive alignment:
\begin{equation}
d_t =
\begin{cases}
\displaystyle \frac{\max(0,r_t^\top u_t)}{\|u_t\|_2^2+\varepsilon}\,u_t, & \|u_t\|_2^2>\delta,\\[6pt]
0, & \text{otherwise},
\end{cases}
\label{eq:positive_projection}
\end{equation}
where $\varepsilon$ stabilizes the denominator and $\delta$ excludes negligible auxiliary responses. The distillation target is then
\begin{equation}
q_t=\operatorname{softmax}\!\left(\log p_{S,t}+d_t\right).
\label{eq:response_target}
\end{equation}
The target adjusts the student distribution along the auxiliary response direction when $\|u_t\|_2^2>\delta$ and $r_t^\top u_t>0$; otherwise, $q_t=p_{S,t}$.

\subsection{Distillation Objective}
\label{sec:distillation_objective}

We train the NC student without value intervention to match the response-guided target $q_t$. We additionally retain a weak CE teacher supervision term, whose weight increases as the relative magnitude of the retained correction decreases:
\begin{equation}
w_t = \operatorname{clip}\!\left(1-\frac{\|d_t\|_2}{\|r_t\|_2+\varepsilon},\,0,\,1\right).
\label{eq:teacher_weight}
\end{equation}
The overall objective is
\begin{equation}
\mathcal{L} = \frac{1}{|\mathcal{I}|} \sum_{t\in\mathcal{I}} \left[\operatorname{JSD}(p_{S,t},q_t) + \lambda w_t\operatorname{JSD}(p_{S,t},p_{T,t})\right],
\label{eq:distillation_loss}
\end{equation}
where $\mathcal{I}$ denotes valid response-token positions in the student-generated trajectories, $\operatorname{JSD}$ is the Jensen--Shannon divergence~\cite{distance1991}, and $\lambda$ controls the strength of CE teacher supervision. Both terms use the shared top-$K$-plus-tail representation. The targets $q_t$, teacher distributions $p_{T,t}$, and weights $w_t$ are detached from gradient computation; only the NC student is updated.
\section{Experiments and Results}
\label{sec:experiments}

\begin{table*}[t]
\centering
\caption{Comparison with general and medical MLLMs and on-policy distillation baselines on internal and external test sets. ``+'' denotes further training from SFT. Gray-shaded Average columns report separate unweighted means across the three internal and external test sets. Best and second-best available results within each Qwen3-VL block are \textbf{bold} and \underline{underlined}, respectively. Double underlines identify the best available off-the-shelf result in each column across general and medical MLLMs. $\Delta$ reports Ours minus the best-performing off-the-shelf MLLM in each column, considering both general and medical models, with \textcolor{gainred}{red} indicating improvement and \textcolor{dropblue}{blue} indicating decline.}
\label{tab:main_comparison}
\small
\setlength{\tabcolsep}{3pt}
\setlength{\aboverulesep}{0.15ex}
\setlength{\belowrulesep}{0.2ex}
\renewcommand{\arraystretch}{0.90}
\begin{tabularx}{\textwidth}{
    lc
    *{6}{>{\centering\arraybackslash}X}
    *{2}{>{\columncolor{gray!8}\centering\arraybackslash}X}
}
\toprule
\multirow{2}{*}{Model / Method} & \multirow{2}{*}{Param.} & \multicolumn{2}{c}{Liver Cancer} & \multicolumn{2}{c}{Brain Cancer} & \multicolumn{2}{c}{Breast Cancer} & \multicolumn{2}{>{\columncolor{gray!8}}c}{Average} \\
\cmidrule(lr){3-4}\cmidrule(lr){5-6}\cmidrule(lr){7-8}\cmidrule(lr){9-10}
& & Internal & External & Internal & External & Internal & External & Internal & External \\
\midrule
\rowcolor{blue!4}\multicolumn{10}{c}{\textit{General Off-the-Shelf MLLMs}} \\
GPT-5.6 Luna~\cite{openai2026gpt56luna} & -- & 50.72 & 44.90 & 46.98 & 47.06 & 47.52 & 47.04 & 48.41 & 46.33 \\
Gemini 3.5 Flash~\cite{google2026gemini35flash} & -- & \bestref{54.26} & 51.82 & \bestref{68.24} & \bestref{67.14} & \bestref{64.01} & \bestref{57.65} & \bestref{62.17} & \bestref{58.87} \\
DeepSeek-V4.1-Flash~\cite{deepseekai2026deepseekv41flashpushinglimitskv} & 552B & 44.54 & 53.58 & 55.25 & 55.41 & 53.05 & 49.63 & 50.95 & 52.87 \\
Qwen3.8-Flash~\cite{qwen2026qwen38flash} & -- & 49.43 & 51.23 & 53.38 & 53.64 & 48.27 & 47.38 & 50.36 & 50.75 \\
Qwen3.8-Max~\cite{qwen2026qwen38max} & 2.4T & 52.60 & \bestref{56.98} & 57.06 & 54.91 & 54.70 & 52.07 & 54.79 & 54.65 \\
Kimi K3~\cite{kimiteam2026kimik3openfrontier} & 2.8T & 49.66 & 55.57 & 53.26 & 53.99 & 52.26 & 52.01 & 51.73 & 53.86 \\
\midrule
\rowcolor{blue!4}\multicolumn{10}{c}{\textit{Medical Off-the-Shelf MLLMs}} \\
Lingshu-7B~\cite{xu2026lingshugeneralistfoundationmodel} & 7B & 42.20 & 43.14 & 47.41 & 47.64 & 50.92 & 47.86 & 46.84 & 46.21 \\
HealthGPT-Pro-4B~\cite{lin2025healthgptmedicallargevisionlanguage} & 4B & 44.08 & 37.51 & 42.63 & 42.98 & 38.86 & 38.48 & 41.86 & 39.66 \\
HuatuoGPT-Vision~\cite{chen2024huatuogptvisioninjectingmedicalvisual} & 7B & 46.50 & 41.74 & 45.15 & 45.99 & 40.61 & 40.72 & 44.09 & 42.82 \\
HealthGPT-Pro-8B~\cite{lin2025healthgptmedicallargevisionlanguage} & 8B & 46.65 & 43.96 & 46.89 & 47.04 & 41.80 & 40.79 & 45.11 & 43.93 \\
Lingshu-32B~\cite{xu2026lingshugeneralistfoundationmodel} & 32B & 37.00 & 38.69 & 46.86 & 48.52 & 46.17 & 47.59 & 43.34 & 44.93 \\
\midrule
\rowcolor{blue!4}\multicolumn{10}{c}{\textit{Qwen3-VL-2B}~\cite{bai2025qwen3}} \\
Base & 2B & 43.41 & 37.51 & 42.91 & 41.00 & 40.33 & 41.54 & 42.22 & 40.02 \\
SFT & 2B & \underline{65.34} & \underline{48.89} & 86.09 & 82.27 & 79.01 & \underline{73.90} & \underline{76.81} & \underline{68.35} \\
+ OPSD~\cite{zhao2026self} & 2B & 63.98 & 47.60 & \underline{86.13} & 82.13 & 79.29 & 73.62 & 76.47 & 67.78 \\
+ Med-OPD~\cite{qian2026med} & 2B & 64.36 & 46.54 & 85.82 & \underline{82.29} & 79.29 & 73.69 & 76.49 & 67.51 \\
+ VA-OPD~\cite{liu2026visual} & 2B & 62.85 & 46.66 & 86.03 & 82.05 & \underline{79.32} & 73.49 & 76.07 & 67.40 \\
\rowcolor{gray!8} \textbf{+ VIVID (Ours)} & 2B & \textbf{66.47} & \textbf{52.64} & \textbf{86.37} & \textbf{82.53} & \textbf{79.44} & \textbf{75.25} & \textbf{77.43} & \textbf{70.14} \\
\rowcolor{gray!8} $\Delta$ (vs.\ Best Off-the-Shelf) & -- & \scorediff{+12.21} & \scorediff{-4.34} & \scorediff{+18.13} & \scorediff{+15.39} & \scorediff{+15.43} & \scorediff{+17.60} & \scorediff{+15.26} & \scorediff{+11.27} \\
\midrule
\rowcolor{blue!4}\multicolumn{10}{c}{\textit{Qwen3-VL-4B}~\cite{bai2025qwen3}} \\
Base & 4B & 43.48 & 39.39 & 44.44 & 44.81 & 44.39 & 42.90 & 44.10 & 42.37 \\
SFT & 4B & 68.50 & 54.75 & 85.94 & 82.90 & 79.44 & 74.71 & 77.96 & 70.79 \\
+ OPSD~\cite{zhao2026self} & 4B & 70.01 & 55.33 & 86.09 & 82.35 & \underline{80.35} & 74.58 & 78.82 & 70.75 \\
+ Med-OPD~\cite{qian2026med} & 4B & 70.69 & 55.45 & \underline{86.34} & \underline{83.00} & 80.29 & 74.58 & 79.11 & 71.01 \\
+ VA-OPD~\cite{liu2026visual} & 4B & \underline{70.99} & \underline{55.80} & \underline{86.34} & 82.43 & 80.19 & \underline{74.92} & \underline{79.17} & \underline{71.05} \\
\rowcolor{gray!8} \textbf{+ VIVID (Ours)} & 4B & \textbf{71.29} & \textbf{57.91} & \textbf{86.46} & \textbf{83.06} & \textbf{80.72} & \textbf{75.19} & \textbf{79.49} & \textbf{72.05} \\
\rowcolor{gray!8} $\Delta$ (vs.\ Best Off-the-Shelf) & -- & \scorediff{+17.03} & \scorediff{+0.93} & \scorediff{+18.22} & \scorediff{+15.92} & \scorediff{+16.71} & \scorediff{+17.54} & \scorediff{+17.32} & \scorediff{+13.18} \\
\bottomrule
\end{tabularx}
\end{table*}

\subsection{Experiment Setup}
\label{sec:experimental_setup}

\noindent\textbf{Datasets.} We evaluate models on our proposed paired diagnostic dataset covering liver CT, brain MRI, and breast MRI, using both internal and external test sets. All methods receive only NC images at inference. We further evaluate zero-shot generalization on VQA-RAD~\cite{lau2018dataset}, SLAKE~\cite{liu2021slake}, PMC-VQA~\cite{zhang2023pmc}, and ChartQA~\cite{masry2022chartqa}.

\noindent\textbf{Evaluation metrics.} We report per-test-set answer accuracy and unweighted averages across the three anatomical sites, separately for internal and external tests. Cross-benchmark evaluation uses only closed-ended questions under each benchmark's protocol, with an unweighted average of benchmark accuracies.

\noindent\textbf{Implementation details.} We fine-tune each backbone for two epochs, then train all on-policy distillation methods for one additional epoch from the same backbone-specific SFT checkpoint. Main comparisons use Qwen3-VL-2B/4B~\cite{bai2025qwen3}, with InternVL3-2B~\cite{zhu2025internvl3} and MedGemma-4B~\cite{sellergren2025medgemma} used to assess applicability across architectures. For \textbf{VIVID}, we set $\alpha=0.5$, $\tau=0.5$, $K=100$, and $\lambda=0.1$. Further training details are provided in the Appendix \ref{app:implementation}.

\subsection{Main Results and Ablations}
\label{sec:main_results}

\noindent\textbf{Performance on the paired diagnostic dataset.} Table~\ref{tab:main_comparison} compares general-purpose and medical MLLMs with post-training baselines. Among off-the-shelf models, Gemini 3.5 Flash achieves the highest internal and external average accuracies, outperforming all evaluated medical MLLMs in both averages. Our 4B model surpasses all evaluated off-the-shelf models on every test set, reaching 79.49\% internally and 72.05\% externally, while our 2B model exceeds the best off-the-shelf accuracy on five of six test sets. Among on-policy distillation methods, \textbf{VIVID} ranks first on all six test sets at both scales. At 2B, it improves the internal and external averages over SFT, whereas OPSD, Med-OPD, and VA-OPD underperform SFT in both averages. At 4B, it exceeds the strongest competing on-policy distillation baseline by 0.32 and 1.00, respectively. Multi-seed stability results and a breakdown of accuracy by question type within the paired diagnostic dataset are provided in Appendix~\ref{app:breakdown}.

\noindent\textbf{Generalization across model architectures.} We further evaluate our method on InternVL3-2B and MedGemma-4B using the liver CT benchmark (Table~\ref{tab:cross_architecture}). Our method improves both internal and external accuracy over SFT and OPSD for both architectures. Together with the Qwen3-VL results, these gains support paired visual-value responses as a broadly applicable source of distillation guidance across distinct MLLM families.

\begin{table}[t]
\centering
\caption{Evaluation across MLLM architectures on liver CT dataset. OPSD and our method further train the SFT model. Best and second-best results are \textbf{bold} and \underline{underlined}, respectively.}
\label{tab:cross_architecture}
\small
\setlength{\tabcolsep}{3pt}
\renewcommand{\arraystretch}{0.92}
\setlength{\aboverulesep}{0.2ex}
\setlength{\belowrulesep}{0.25ex}
\begin{tabularx}{\columnwidth}{l*{4}{>{\centering\arraybackslash}X}}
\toprule
\multicolumn{1}{c}{\multirow{2}{*}{Method}} & \multicolumn{2}{c}{\textit{InternVL3-2B}~\cite{zhu2025internvl3}} & \multicolumn{2}{c}{\textit{MedGemma-4B}~\cite{sellergren2025medgemma}} \\
\cmidrule(lr){2-3}\cmidrule(lr){4-5}
& Internal & External & Internal & External \\
\midrule
Zero-shot & 42.65 & 34.82 & 43.86 & 37.16 \\
SFT & \underline{65.26} & 52.17 & \underline{70.61} & \underline{54.51} \\
+ OPSD & 64.81 & \underline{53.22} & 66.84 & 52.40 \\
\rowcolor{gray!8}  \textbf{+ VIVID} & \textbf{66.54} & \textbf{54.04} & \textbf{70.76} & \textbf{59.91} \\
\bottomrule
\end{tabularx}
\end{table}

\begin{table}[t]
\centering
\caption{Cross-benchmark generalization on medical and general VQAs. OPSD and our method further train the SFT model. Avg.\ denotes the unweighted mean accuracy across four benchmarks. Best and second-best results are \textbf{bold} and \underline{underlined}, respectively.}
\label{tab:cross_benchmark}
\small
\setlength{\tabcolsep}{3pt}
\renewcommand{\arraystretch}{0.92}
\setlength{\aboverulesep}{0.2ex}
\setlength{\belowrulesep}{0.25ex}
\begin{tabular*}{\columnwidth}{
  @{\extracolsep{\fill}}
  lcccc>{\columncolor{gray!8}[\tabcolsep][0pt]}c@{}
}
\toprule
Method & VQA-RAD & SLAKE & PMC-VQA & ChartQA & \makebox[0.13\columnwidth][c]{Average} \\
\midrule
\rowcolor{blue!4}\multicolumn{6}{c}{\textit{Qwen3-VL-2B}} \\
SFT & 62.50 & \underline{68.75} & \underline{47.10} & 76.88 & \underline{63.81} \\
+ OPSD & \underline{62.87} & 68.27 & 46.75 & \underline{76.96} & 63.71 \\
\rowcolor{gray!8}  \textbf{+ VIVID} & \textbf{63.24} & \textbf{68.99} & \textbf{47.50} & \textbf{78.56} & \textbf{64.57} \\
\midrule
\rowcolor{blue!4}\multicolumn{6}{c}{\textit{Qwen3-VL-4B}} \\
SFT & 63.97 & \underline{74.04} & \underline{51.25} & 81.92 & 67.80 \\
+ OPSD & \underline{64.34} & \textbf{74.76} & 50.85 & \underline{81.96} & \underline{67.98} \\
\rowcolor{gray!8}  \textbf{+ VIVID} & \textbf{65.07} & 73.80 & \textbf{51.80} & \textbf{84.20} & \textbf{68.72} \\
\bottomrule
\end{tabular*}
\end{table}

\noindent\textbf{Cross-task and cross-modality generalization.} We evaluate the trained models on four additional VQA benchmarks spanning tasks and image modalities outside our post-training data, without further adaptation (Table~\ref{tab:cross_benchmark}). Our method achieves the highest average score at both model scales, outperforming SFT and OPSD on all four benchmarks at 2B and three of four at 4B. These gains suggest that paired medical supervision strengthens transferable visual question-answering capabilities beyond the specific tasks and acquisition conditions from which it is derived.

\noindent\textbf{Ablation Studies.} We evaluate three variants on the liver CT dataset at both 2B and 4B scales (Table~\ref{tab:ablation}): (1) \textit{Random Reg.}, which replaces foreground attention-guided selection with an equal number of randomly sampled image positions; (2) \textit{Alt. Attn.}, which constructs the auxiliary branch by mixing paired spatial attention weights instead of visual values; and (3) \textit{w/o Proj.}, which directly distills the auxiliary prediction. The full method performs best across all internal and external settings. Additional ablation studies, including efficiency analysis and sensitivity to other hyperparameters, are provided in Appendix \ref{app:moreablation}.

\begin{table}[t]
\centering
\caption{Ablation study across Qwen3-VL on the liver CT dataset. Best and second-best results are \textbf{bold} and \underline{underlined}, respectively.}
\label{tab:ablation}
\small
\setlength{\tabcolsep}{3pt}
\renewcommand{\arraystretch}{0.92}
\setlength{\aboverulesep}{0.2ex}
\setlength{\belowrulesep}{0.25ex}
\begin{tabularx}{\columnwidth}{l*{4}{>{\centering\arraybackslash}X}}
\toprule
\multicolumn{1}{c}{\multirow{2}{*}{Variant}} & \multicolumn{2}{c}{\textit{Qwen3-VL-2B}} & \multicolumn{2}{c}{\textit{Qwen3-VL-4B}} \\
\cmidrule(lr){2-3}\cmidrule(lr){4-5}
& Internal & External & Internal & External \\
\midrule
Random Reg. & 65.71 & 49.82 & \underline{71.14} & 56.51 \\
Alt. Attn. & 64.73 & 51.47 & 71.06 & 54.28 \\
w/o Proj. & \underline{66.24} & \underline{51.70} & 70.99 & \underline{57.33} \\
\textbf{VIVID} & \textbf{66.47} & \textbf{52.64} & \textbf{71.29} & \textbf{57.91} \\
\bottomrule
\end{tabularx}
\end{table}

\subsection{Analysis of Distributions and Visual Values}
\label{sec:analysis}

We compare SFT, OPSD, and our method on the liver CT dataset using Qwen3-VL-2B and 4B, examining both predictive distributions and internal visual representations.

\noindent\textbf{Task-critical probability allocation.} We define \emph{margin selectivity} (MS) as the difference in average token margin between medical-judgment spans and the remaining evidence tokens, where token margin is the top-1--top-2 probability gap. We additionally measure correct-option NLL over the full evaluation set. On questions answered incorrectly by all three models, we report the mean probability assigned to the correct option ($P_{\mathrm{GT}}^{\mathrm{err}}$). Our method achieves the best results on all three metrics at both scales (Table~\ref{tab:predictive_analysis}). These results reveal more selective confidence at task-critical positions alongside stronger support for correct answers, including those that remain unselected.

\begin{figure}[t]
\centering
\includegraphics[width=\columnwidth]{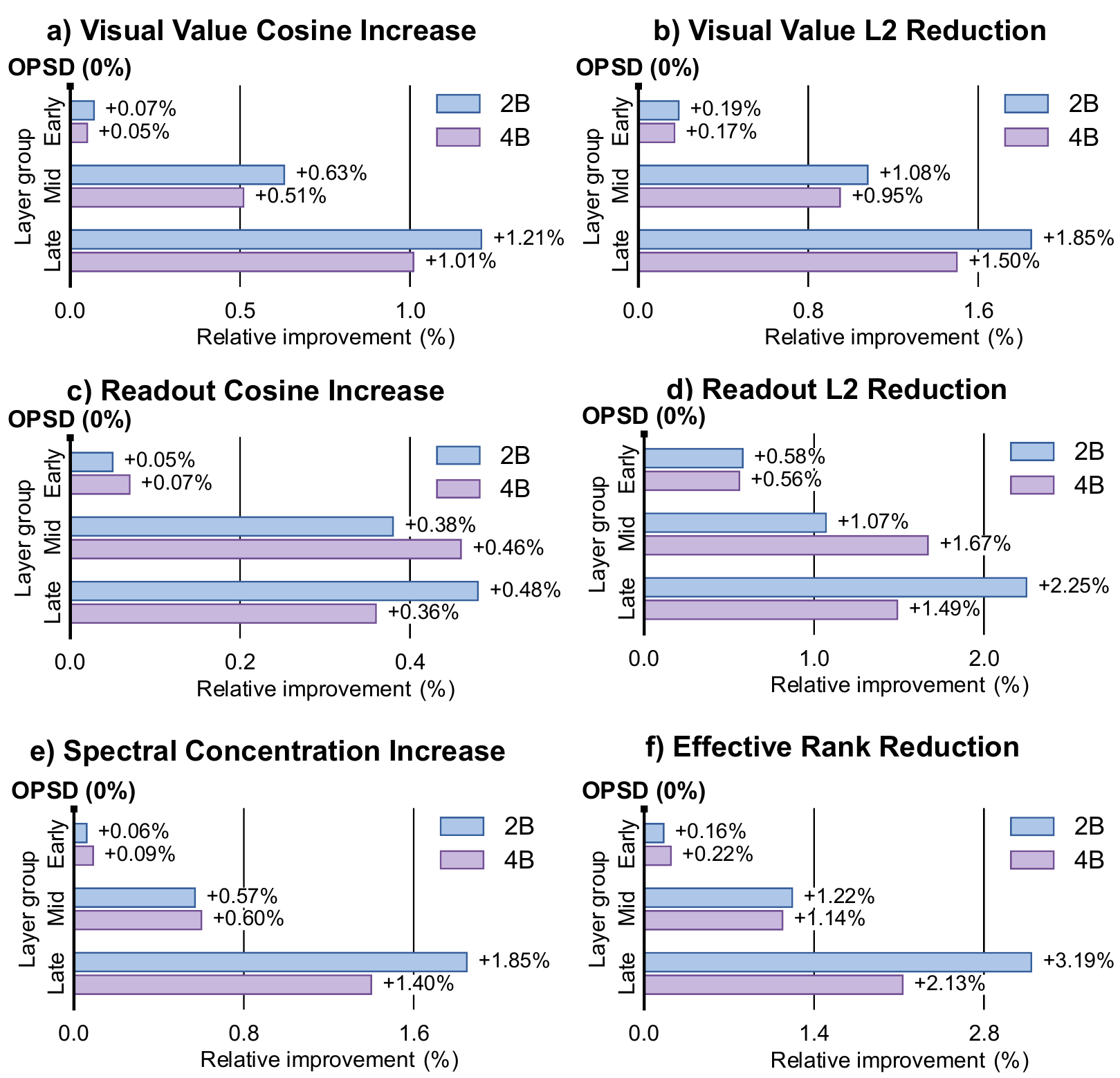}
\caption{\textbf{Visual representation improvement relative to OPSD on liver CT dataset.} Blue and purple denote Qwen3-VL-2B and 4B. Percentage changes relative to OPSD are grouped by decoder depth: (a--d) increased cosine similarity and reduced $\ell_2$ distance to the enhanced-image reference for visual values and fixed-attention readouts; (e) increased top-ten explained variance; (f) reduced effective rank. Reductions are plotted as positive values.}
\label{fig:explain}
\end{figure}

\noindent\textbf{Visual representation changes.} We assess alignment with the frozen SFT model's enhanced-image visual values using cosine similarity and relative $\ell_2$ distance over foreground visual values $V$ and task-critical readouts $MV$. For the attention readout $MV$ comparison, we apply identical SFT-NC attention weights to all branches. Relative to OPSD, our method improves both alignment measures across all three depth groups, with larger changes generally beyond the early layers (Fig.~\ref{fig:explain}a--d). Visual values also exhibit greater top-ten explained variance and lower effective rank, indicating a more concentrated spectrum (Fig.~\ref{fig:explain}e,f). Thus, distribution-level distillation is accompanied by closer alignment with enhanced visual representations and a more compact spectral structure, without explicit feature-alignment or spectral regularization. Additional analyzes, case studies, and evaluation of generated evidence are provided in Appendices \ref{app:intervention_behavior} and \ref{app:expert_evaluation}.

\begin{table}[t]
\centering
\caption{Predictive-distribution analysis on liver CT dataset. Best and second-best values within each model scale are \textbf{bold} and \underline{underlined}, respectively.}
\label{tab:predictive_analysis}
\small
\setlength{\tabcolsep}{3pt}
\renewcommand{\arraystretch}{0.92}
\setlength{\aboverulesep}{0.2ex}
\setlength{\belowrulesep}{0.25ex}
\begin{tabularx}{\columnwidth}{l*{6}{>{\centering\arraybackslash}X}}
\toprule
\multicolumn{1}{c}{\multirow{2}{*}{Metric}} & \multicolumn{3}{c}{\textit{Qwen3-VL-2B}} & \multicolumn{3}{c}{\textit{Qwen3-VL-4B}} \\
\cmidrule(lr){2-4}\cmidrule(lr){5-7}
& SFT & OPSD & VIVID & SFT & OPSD & VIVID \\
\midrule
MS(pp)$\uparrow$ & 4.92 & \underline{4.99} & \textbf{6.64} & 1.21 & \underline{1.22} & \textbf{2.45} \\
NLL$\downarrow$ & \underline{1.53} & 1.64 & \textbf{1.34} & \underline{0.87} & 0.91 & \textbf{0.85} \\
$P_{\mathrm{GT}}^{\mathrm{err}}(\%)\uparrow$ & \underline{6.45} & 6.23 & \textbf{8.21} & \underline{13.89} & 12.09 & \textbf{14.75} \\
\bottomrule
\end{tabularx}
\end{table}

\section{Conclusion}
\label{sec:conclusion}

We revisit visual understanding in MLLMs beyond spatial attention maps. Using paired NC and CE images, we show that similar attention patterns can lead to different predictions, while controlled interventions reveal that visual values exert greater influence on answer correctness than attention weights. Accordingly, VIVID uses the student's response to enhanced visual values to refine teacher supervision for on-policy distillation. Experiments show consistent gains across anatomical sites, modalities, and model architectures. Our findings stress not only where MLLMs attend, but also what information they retrieve, offering a perspective on visual understanding and knowledge transfer.

\section{Acknowledgment}
Supported by National Natural Science Foundation of China under Grant 62271465, National Key R\&D Program of China under Grant 2025YFC3408300,  Basic Research Program of Jiangsu (No. BK20255001), Suzhou Basic Research Program under Grant SYG202338, and the Sanming Project of Medicine in Shenzhen under Grant SZSM202411033.




\clearpage
\maketitlesupplementary
\appendix

\startcontents[appendix]

\section*{Supplementary Contents}

\printcontents[appendix]{}{1}{\setcounter{tocdepth}{2}}


\begin{table}[t]
\centering
\caption{\textbf{Patient counts by source cohort and split.} Int. and Ext. denote internal and external, respectively.}
\label{tab:source_cohort_splits}
\small
\setlength{\tabcolsep}{2pt}
\renewcommand{\arraystretch}{1.05}
\begin{tabular*}{\columnwidth}{@{\extracolsep{\fill}}lcccc@{}}
\toprule
Source cohort & Train & Int. Test & Ext. Test & Total \\
\midrule
\rowcolor{blue!4}\multicolumn{5}{c}{\textit{Liver CT}} \\
MCT-LTDiag & 450 & 50 & 0 & 500 \\
TriALS & 0 & 0 & 44 & 44 \\
\midrule
\rowcolor{blue!4}\multicolumn{5}{c}{\textit{Brain MRI}} \\
UPENN-GBM & 361 & 42 & 0 & 403 \\
UCSF-PDGM & 242 & 21 & 0 & 263 \\
CPTAC-GBM & 26 & 7 & 0 & 33 \\
IvyGAP & 28 & 2 & 0 & 30 \\
ACRIN-FMISO & 4 & 0 & 0 & 4 \\
Additional cases\textsuperscript{a} & 315 & 36 & 0 & 351 \\
TCGA-GBM & 0 & 0 & 102 & 102 \\
TCGA-LGG & 0 & 0 & 65 & 65 \\
\midrule
\rowcolor{blue!4}\multicolumn{5}{c}{\textit{Breast MRI}} \\
Duke & 268 & 23 & 0 & 291 \\
I-SPY1 & 153 & 18 & 0 & 171 \\
I-SPY2 & 871 & 103 & 0 & 974 \\
NACT & 0 & 0 & 64 & 64 \\
\midrule
Total & 2,718 & 302 & 275 & 3,295 \\
\bottomrule
\end{tabular*}
\par\vspace{3pt}
\begin{minipage}{\columnwidth}
\footnotesize
\textsuperscript{a}Additional BraTS 2021 cases listed under
new-not-previously-in-TCIA in the source metadata.
\end{minipage}
\end{table}

\section{Cohort Composition and Patient-Level Partitioning}
\label{app:benchmark}

\subsection{Data Sources and Patient Splits}
\label{app:data_sources}

Table~\ref{tab:source_cohort_splits} summarizes the source cohorts and patient-level splits for the three datasets. All splits are patient-disjoint.

\noindent\textbf{Liver CT.} We use MCT-LTDiag~\cite{wu2025multi} for training and internal evaluation, randomly dividing its 500 patients into 450 training and 50 internal-test patients. An additional 44 patients from TriALS~\cite{elbatel2026trials} are reserved for external testing. We pair non-contrast scans with portal venous phase (PVP) images as the contrast-enhanced view. TriALS contributes only NC images for external evaluation.

\noindent\textbf{Brain MRI.} We use paired native and contrast-enhanced T1-weighted scans. Images and segmentation annotations are obtained through BraTS 2021~\cite{baid2021rsna}. The development pool comprises 1,084 patients from UPENN-GBM~\cite{bakas2022university}, UCSF-PDGM~\cite{calabrese2022university}, CPTAC-GBM~\cite{national2023clinical}, IvyGAP~\cite{puchalski2018anatomic}, ACRIN-FMISO-Brain~\cite{kinahan2018acrin}, and additional cases contributed to the BraTS 2021 release. These patients are randomly divided into 976 training and 108 internal-test patients. The 167 patients from TCGA-GBM~\cite{scarpace2016tcgagbm} and TCGA-LGG~\cite{pedano2016tcgalgg} are reserved for external testing.

\noindent\textbf{Breast MRI.} We pair pre-contrast scans with the first post-contrast acquisition, using images and segmentation annotations obtained through MAMA-MIA~\cite{garrucho2025large}. The 1,436 patients from Duke~\cite{saha2018machine}, I-SPY1~\cite{newitt2016ispy1}, and I-SPY2~\cite{li2022ispy2,newitt2021acrin} are randomly divided into 1,292 training and 144 internal-test patients. All 64 patients from NACT~\cite{newitt2016nact} are reserved for external testing.

\begin{figure*}[!htbp]
\centering
\includegraphics[width=\textwidth]{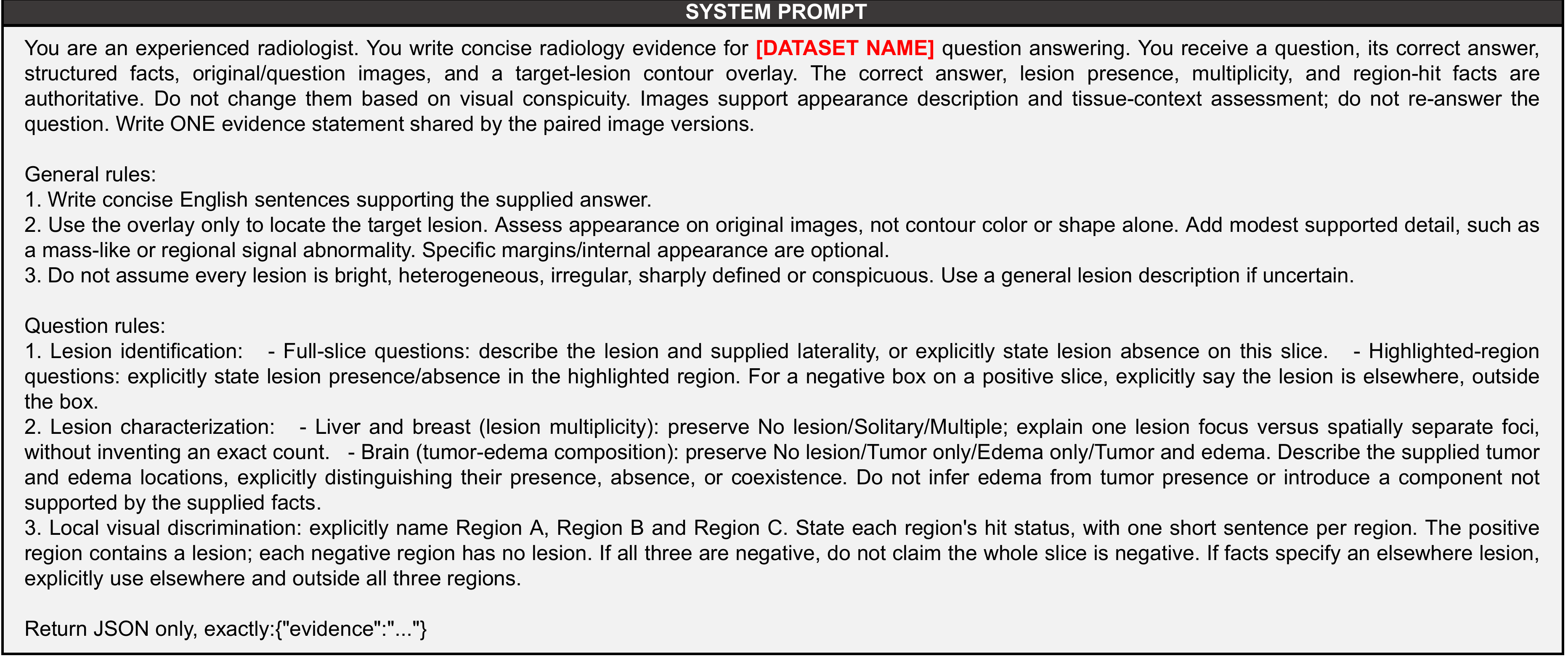}
\caption{\textbf{Prompt template for evidence generation.} \texttt{[DATASET NAME]} is replaced with the corresponding dataset name.}
\label{fig:prompt}
\end{figure*}

\subsection{Image Preprocessing and Standardization}
\label{app:image_preparation}

\noindent\textbf{Image Pairing and Spatial Consistency.} For liver CT, we pair non-contrast and portal venous phase (PVP) images from the same patient. PVP is selected for its diagnostic value in detecting hepatic lesions~\cite{soyer2004detection}. For brain MRI, we pair native T1-weighted and contrast-enhanced T1-weighted images, consistent with recommendations to acquire pre- and post-contrast T1-weighted sequences in standardized brain tumor imaging protocols~\cite{ellingson2015consensus}. For breast MRI, we pair the pre-contrast acquisition with the first post-contrast acquisition, which is supported by clinical studies demonstrating the utility of early post-contrast images for breast cancer detection~\cite{kuhl2014abbreviated}. For paired data, corresponding slices are extracted on the associated image grids, with consistent spatial markers across views. Image--annotation grid consistency is checked during preparation. Since the source TriALS scans are not registered across contrast phases, we use only the NC images without constructing NC--CE pairs; this does not affect its role as an external test set, as our external evaluation requires only NC inputs.

\noindent\textbf{Modality-Specific Intensity Normalization.} CT images are processed using fixed phase-specific liver windows: a window level of 45 HU and a width of 120 HU for NC images, and a level of 85 HU with the same width for CE images. The higher CE window level accommodates the increased attenuation following contrast administration. Our use of the selected liver width and liver window is motivated by studies examining the benefits of dedicated liver-window settings for lesion detection~\cite{mayosmith1999detecting,okimoto2024new}. Intensities are clipped to the corresponding window and linearly mapped to $[0,255]$. MRI sequences are normalized independently at the volume level. The 0.5th and 99.5th percentiles of finite, nonzero voxel intensities define the clipping bounds, after which intensities are linearly mapped to $[0,255]$ while zero-valued background remains zero. All slices from a volume share the same normalization parameters. For breast MRI, these percentiles are estimated using deterministic voxel subsampling.

\noindent\textbf{Spatial Resampling and Image Export.} All images are exported in PNG format at $512\times512$ pixels. Liver CT slices are center-cropped to $512\times512$. Brain MRI slices are axial and upsampled from $240\times240$ to $512\times512$ using bicubic interpolation. Breast MRI retains the source acquisition plane: axial for Duke and I-SPY2, and sagittal for I-SPY1 and NACT. Breast slices are first resized according to their physical dimensions using bilinear interpolation and centered zero-padding to form $384\times384$ images, then upsampled to $512\times512$ using bicubic interpolation. Segmentation masks use nearest-neighbor interpolation, and spatial annotations are transformed consistently with the images.

\subsection{Annotation-Guided Sampling and Question Construction}
\label{app:slice_question_construction}

\noindent\textbf{Lesion-guided slice sampling.}
For each patient, we select 3--6 representative lesion-positive slices, prioritizing substantial lesion coverage and diversity in the available lesion-attribute categories. We additionally sample up to two negative slices that contain sufficient anatomical tissue but no lesion for hard negative question construction. To limit redundancy among neighboring slices, we enforce a dataset-specific minimum separation in slice indices. This separation is progressively relaxed when too few candidates are available; if the desired sample count still cannot be reached, we retain fewer slices.

\noindent\textbf{Question types.}
We construct three question types spanning slice-level assessment and region-level discrimination, motivated by the combination of global assessment and focal examination in radiological interpretation~\cite{kundel2007holistic}.
\textit{Lesion identification} asks whether a target lesion is present within the queried area, which is either the full slice or a highlighted region.
\textit{Lesion characterization} assesses whether lesions are solitary or multiple in liver and breast imaging, or the tumor--edema composition in brain imaging.
\textit{Local visual discrimination} asks which of three candidate regions contains a target lesion, with an additional ``None of the above'' option.
Questions and reference answers are derived from segmentation annotations. Region-based questions are omitted when valid candidate regions cannot be constructed.

\begin{figure*}[!htbp]
\centering
\includegraphics[width=\textwidth]{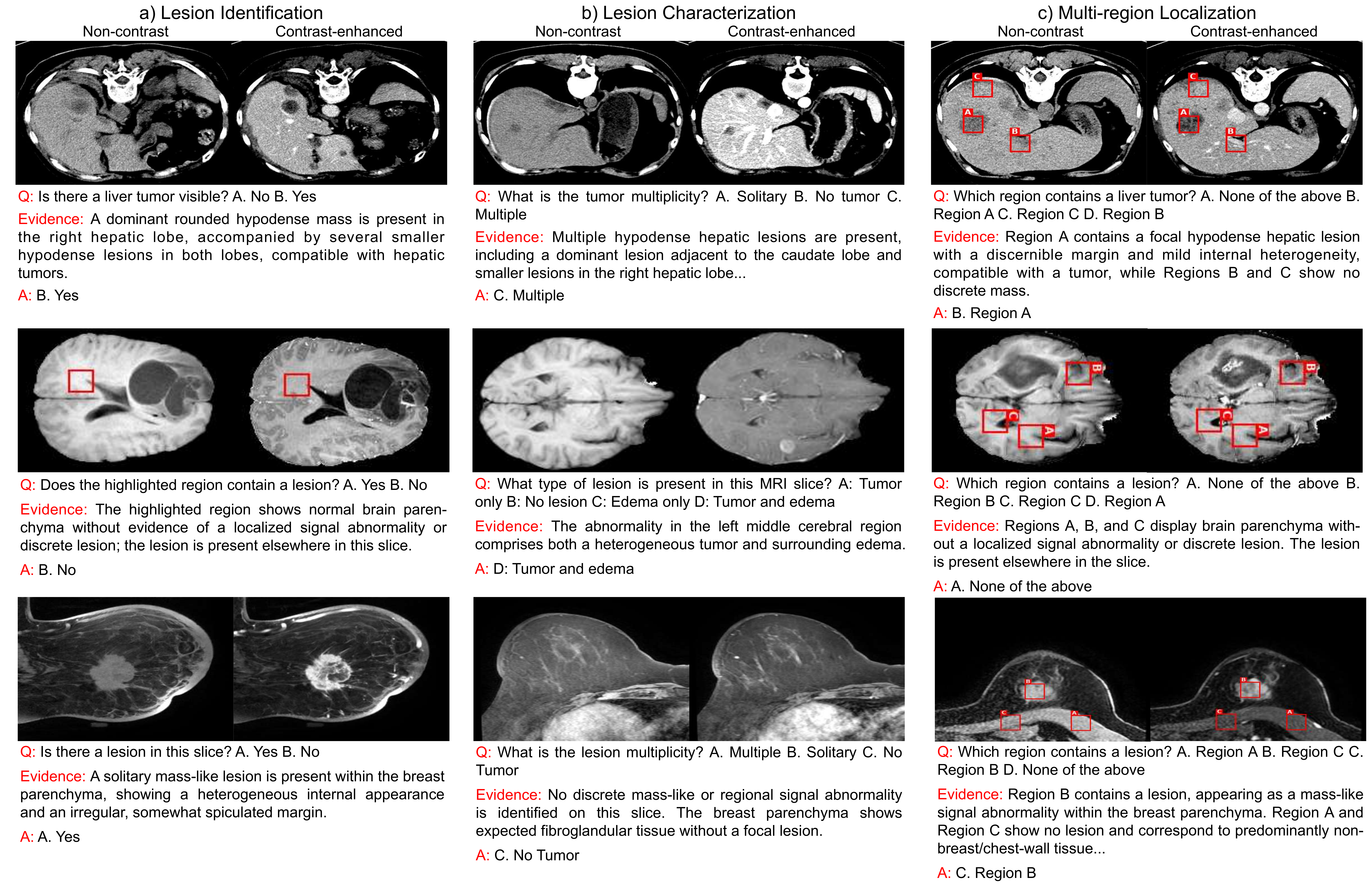}
\caption{\textbf{Representative examples of our dataset.} From left to right: lesion identification, lesion characterization, and multi-region localization.}
\label{fig:dataset_cases}
\end{figure*}

\noindent\textbf{Shortcut mitigation.}
Clinical image interpretation requires distinguishing a suspected lesion from an actual finding: a region selected for closer inspection does not necessarily contain a lesion. Similarly, a lesion elsewhere on the slice does not establish lesion presence within the region under assessment. Motivated by these distinctions, we construct region-based questions that require verifying lesion presence within the specified spatial scope. This design also addresses the tendency of medical VQA models to rely on textual associations or irrelevant image regions rather than question-specific visual evidence~\cite{nguyen2025localizing}. Specifically, we include lesion-negative boxes on lesion-positive slices, requiring models to assess the queried region independently of findings elsewhere. Regional negatives also include candidates mined using NC appearance features. Lesion-characterization questions retain an explicit ``No lesion'' option, requiring models to establish lesion presence before assessing multiplicity or tumor--edema composition. Likewise, local visual discrimination questions may present three lesion-negative regions even when a lesion exists elsewhere on the slice, making ``None of the above'' the correct answer. These cases are designed to discourage treating a highlighted region, a lesion-related question, or an abnormal slice as sufficient evidence of a positive finding.

\begin{table}[!htbp]
\centering
\caption{Five-point rubric for spatial consistency. Ratings assess anatomical location, laterality, region identifiers, and inside/outside relations against the supplied annotations.}
\label{tab:spatial_rubric}
\small
\setlength{\tabcolsep}{6pt}
\renewcommand{\arraystretch}{1.12}
\begin{tabularx}{\columnwidth}{@{}c >{\raggedright\arraybackslash}X@{}}
\toprule
\textbf{Score} & \textbf{Assessment criteria} \\
\midrule
5 & All spatial references are correct and clearly identify the location or region relevant to the question. \\
\addlinespace[5pt]
4 & Spatial references are correct and unambiguous, with only minor wording imprecision that does not affect localization. \\
\addlinespace[5pt]
3 & The intended location remains identifiable, but localized errors, omissions, or ambiguous references require substantive clarification. \\
\addlinespace[5pt]
2 & A major error in anatomical location, laterality, region identity, or inside/outside assignment prevents reliable localization, although some spatial details remain correct. \\
\addlinespace[5pt]
1 & Spatial references predominantly identify the wrong site or region, contradict the annotations, or fail to establish an interpretable location. \\
\bottomrule
\end{tabularx}
\end{table}

\begin{table}[!htbp]
\centering
\caption{Five-point rubric for factual accuracy and answer support. Ratings assess medical correctness, grounding in the supplied facts and images, and support for the reference answer.}
\label{tab:factual_rubric}
\small
\setlength{\tabcolsep}{6pt}
\renewcommand{\arraystretch}{1.12}
\begin{tabularx}{\columnwidth}{@{}c >{\raggedright\arraybackslash}X@{}}
\toprule
\textbf{Score} & \textbf{Assessment criteria} \\
\midrule
5 & All claims are medically accurate, supported by the supplied information, and relevant to the question. The evidence clearly supports the reference answer without unsupported findings or unjustified certainty. \\
\addlinespace[5pt]
4 & The evidence accurately supports the reference answer, with only minor wording imprecision or redundancy and no substantive factual errors. \\
\addlinespace[5pt]
3 & Non-critical factual errors, omissions, or insufficient explanation weaken answer support, but the core finding remains correct. \\
\addlinespace[5pt]
2 & Major errors or unsupported inferences concerning lesion presence, multiplicity, composition, or appearance undermine answer support, although some valid information remains. \\
\addlinespace[5pt]
1 & The central description contradicts the supplied facts or reference answer, reverses the key finding, or consists predominantly of unsupported claims. \\
\bottomrule
\end{tabularx}
\end{table}

\begin{table}[!htbp]
\centering
\caption{\textbf{Dataset statistics.} Patient / question counts for training, internal testing, and external testing. Each question is counted once across paired NC--CE views.}
\label{tab:dataset_statistics}
\small
\setlength{\tabcolsep}{1.5pt}
\renewcommand{\arraystretch}{1.1}
\begin{tabular*}{\columnwidth}{@{\extracolsep{\fill}}lcccc@{}}
\toprule
Dataset & Train & Int. test & Ext. test & Total \\
\midrule
Liver  & 450 / 11778  & 50 / 1327  & 44 / 853   & 544 / 13958  \\
Brain  & 976 / 29282  & 108 / 3265 & 167 / 4981 & 1251 / 37528 \\
Breast & 1292 / 28840 & 144 / 3201 & 64 / 1471  & 1500 / 33512 \\
\midrule
Total  & 2718 / 69900 & 302 / 7793 & 275 / 7305 & 3295 / 84998 \\
\bottomrule
\end{tabular*}
\end{table}

\subsection{Annotation-Grounded Evidence Generation}
\label{app:evidence_generation}

\noindent\textbf{Factual constraints from annotations.}
We derive factual constraints from segmentation annotations, including lesion presence, task-specific lesion attributes, and spatial relationships between lesions and candidate regions. For liver CT, we additionally obtain liver-segment masks using TotalSegmentator~\cite{Wasserthal_2023} and assign lesions to anatomical segments based on spatial overlap. These assignments provide anatomical references for evidence descriptions.

\noindent\textbf{Constrained generation and consistency checks.}
We use Gemini 3.1 Pro~\cite{google2026gemini31pro} to generate evidence conditioned on the question, answer options, reference answer, and annotation-derived factual constraints. Each paired question receives a single evidence statement shared across the NC and CE views, followed by the reference answer in the format \texttt{Evidence: ... Answer: ...}. The evidence describes answer-relevant lesion locations, attributes, and relationships to the queried regions, without comparing contrast conditions or claiming equal lesion conspicuity across views. Generation instructions prohibit unsupported diagnoses and findings. Generated statements are checked for consistency with the reference answer's polarity and the region references in the question.

\subsection{Clinical Review and Quality Assurance}
\label{app:clinical_review}

\noindent\textbf{Evaluation rubric.}
A radiologist assesses the generated evidence along two dimensions using five-point Likert scales: \textit{spatial consistency}, which measures agreement with anatomical locations and region annotations, and \textit{factual accuracy and answer support}, which measures whether the evidence is medically accurate and supports the reference answer. Table~\ref{tab:spatial_rubric} and Table~\ref{tab:factual_rubric} provide the scoring criteria.

\noindent\textbf{Pilot assessment and prompt refinement.}
Before full-scale generation, we generate 100 samples for each of the three question types in each dataset. The radiologist reviews all pilot samples, assigning mean scores of 4.3/5.0 for spatial consistency and 4.1/5.0 for factual accuracy and answer support. The primary issue is unsupported inference by the evidence-generating LLM, particularly descriptions of lesion characteristics not established by the supplied information. Based on this review, we revise the generation instructions to constrain evidence to the supplied facts, reference answers, and images, as is provided in Fig.~\ref{fig:prompt}. The revised instructions explicitly prohibit unsupported findings and require general descriptions when specific visual characteristics are uncertain. They also enforce the scope of the queried region, preventing lesion absence within a marked region from being generalized to the entire image.

\noindent\textbf{Post-generation quality assessment.}
Following full-scale generation with the revised instructions, the same radiologist reviews a 20\% sample of the generated dataset using the same rubric. Mean scores are 4.6/5.0 for spatial consistency and 4.5/5.0 for factual accuracy and answer support. These scores indicate that the reviewed evidence is generally spatially consistent, medically accurate, and supportive of the reference answers.

\subsection{Dataset Statistics and Illustrative Examples}
\label{app:dataset_cases}

Table~\ref{tab:dataset_statistics} summarizes the patient and question counts for the three datasets across training, internal testing, and external testing. In total, the benchmark contains 84,998 questions from 3,295 patients, covering liver CT, brain MRI, and breast MRI. The training, internal-test, and external-test sets contain 69,900, 7,793, and 7,305 questions, respectively.

Fig.~\ref{fig:dataset_cases} presents representative examples of lesion identification, lesion characterization, and multi-region localization across the three datasets. Each example includes paired NC and CE images, the question and answer options, and the shared reference evidence and answer. The examples also include lesion-negative queried regions on lesion-positive slices, illustrating why answers must reflect the specified region rather than lesion presence elsewhere in the image.

\section{Additional Analyses of Visual Understanding}
\label{app:attention}

\subsection{Attention Map Similarity Across Datasets}
\label{app:attention_similarity}

\begin{figure}[t]
\centering
\includegraphics[width=\columnwidth]{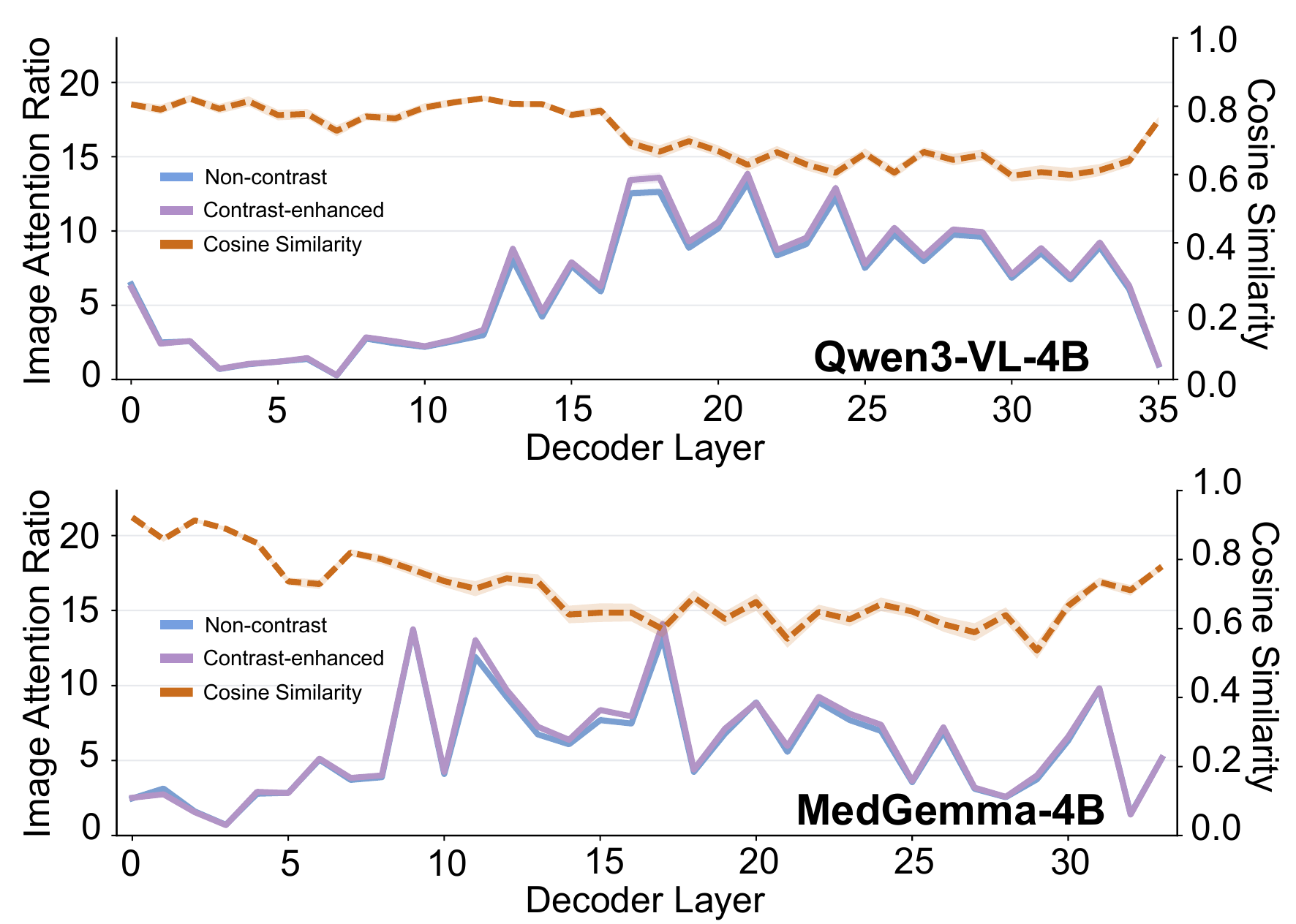}
\caption{\textbf{Layer-wise attention patterns on the brain MRI dataset.} Solid curves show NC and CE image-attention ratios (left axis); dashed curves show foreground cosine similarity (right axis). Shading denotes 95\% confidence intervals.}
\label{fig:attention_brats}
\end{figure}

\begin{figure}[t]
\centering
\includegraphics[width=\columnwidth]{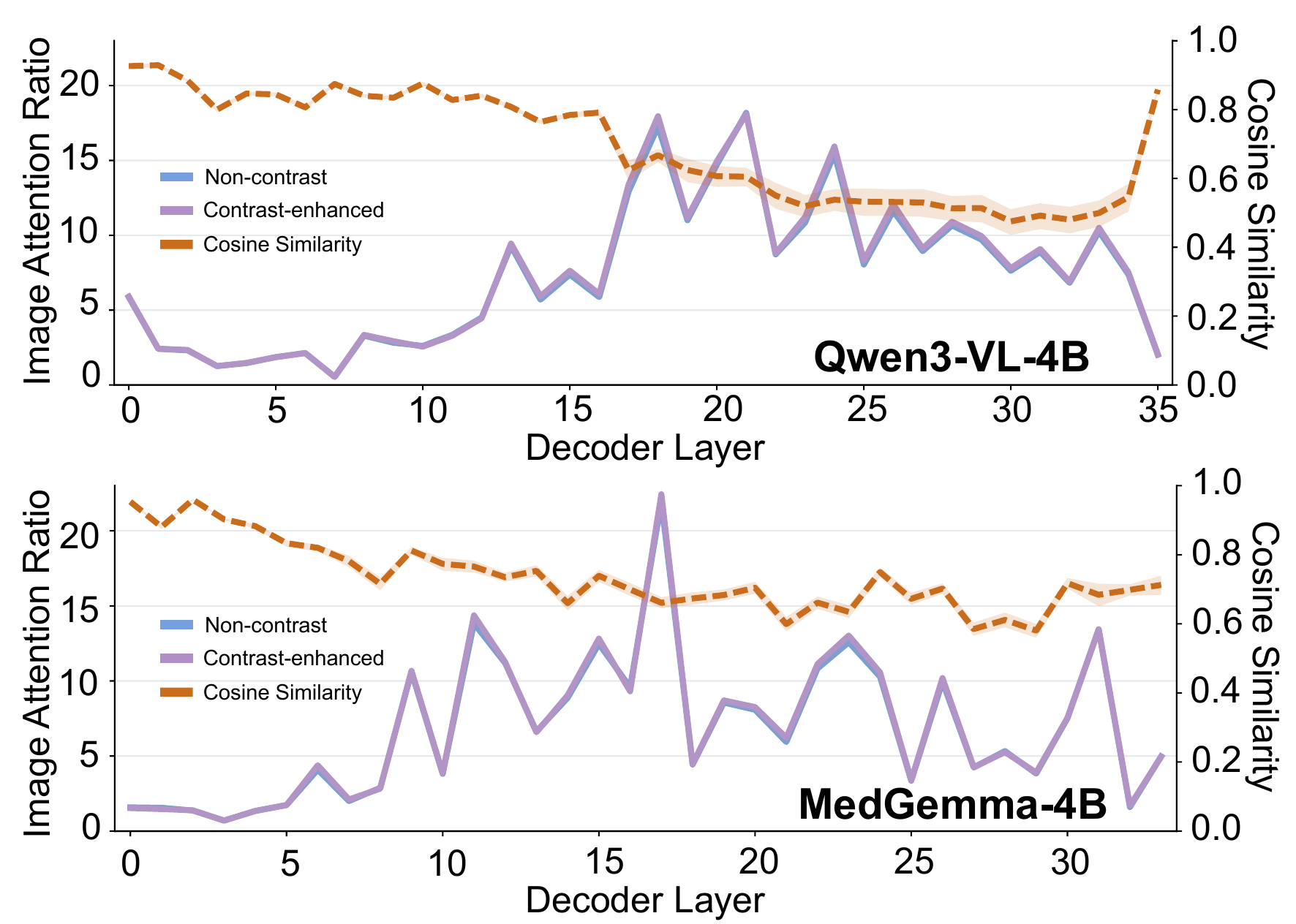}
\caption{\textbf{Layer-wise attention patterns on the breast MRI dataset.} Curves and confidence intervals follow Figure~\ref{fig:attention_brats}}
\label{fig:attention_mamamia}
\end{figure}

\begin{table}[t]
\centering
\caption{\textbf{Paired \texttt{[M]} and \texttt{[V]} exchanges with Qwen3-VL-4B on the reverse subset.} Results are evaluated on NC-correct/CE-incorrect pairs. All-layer changes are reported in percentage points. Arrows indicate donor-to-recipient transfer; red $\Delta$ entries report \texttt{[V]} minus \texttt{[M]}.}
\label{tab:qwen_reverse_exchange}
\small
\setlength{\tabcolsep}{2pt}
\renewcommand{\arraystretch}{1.05}
\setlength{\aboverulesep}{0.2ex}
\setlength{\belowrulesep}{0.25ex}
\begin{tabular*}{\columnwidth}{@{\extracolsep{\fill}}lccccc@{}}
\toprule
\multirow{2}{*}{Dataset}
& \multirow{2}{*}{Swap}
& \multicolumn{2}{c}{$\Delta P_{\mathrm{GT}}$}
& \multicolumn{2}{c}{$\Delta$Acc.} \\
\cmidrule(lr){3-4}\cmidrule(lr){5-6}
& & CE$\to$NC & NC$\to$CE & CE$\to$NC & NC$\to$CE \\
\midrule
\multirow{3}{*}{Liver CT}
& \texttt{[M]} & -0.55 & +1.92 & -26.47 & +19.12 \\
& \texttt{[V]}
& \textbf{-4.51} & \textbf{+5.58}
& \textbf{-42.65} & \textbf{+36.76} \\
& $\Delta$
& \textcolor{gainred}{-3.96} & \textcolor{gainred}{+3.66}
& \textcolor{gainred}{-16.18} & \textcolor{gainred}{+17.65} \\
\midrule
\multirow{3}{*}{Brain MRI}
& \texttt{[M]} & -1.00 & +0.96 & -13.96 & +20.05 \\
& \texttt{[V]}
& \textbf{-2.69} & \textbf{+4.36}
& \textbf{-19.29} & \textbf{+25.38} \\
& $\Delta$
& \textcolor{gainred}{-1.69} & \textcolor{gainred}{+3.41}
& \textcolor{gainred}{-5.33} & \textcolor{gainred}{+5.33} \\
\midrule
\multirow{3}{*}{Breast MRI}
& \texttt{[M]} & -1.10 & +2.66 & -13.33 & +23.70 \\
& \texttt{[V]}
& \textbf{-3.43} & \textbf{+4.69}
& \textbf{-29.63} & \textbf{+40.00} \\
& $\Delta$
& \textcolor{gainred}{-2.32} & \textcolor{gainred}{+2.04}
& \textcolor{gainred}{-16.30} & \textcolor{gainred}{+16.30} \\
\bottomrule
\end{tabular*}
\end{table}

\begin{table}[t]
\centering
\caption{\textbf{Value transfer under attention noise with Qwen3-VL-4B on the reverse subset.} Evaluation subsets follow Table~\ref{tab:qwen_reverse_exchange}. Interventions affect all decoder layers in the NC branch, with noise added only to \texttt{[M]}. Changes are reported in percentage points. Bold entries mark value transfer without attention noise.}
\label{tab:qwen_reverse_noise}
\small
\setlength{\tabcolsep}{2pt}
\renewcommand{\arraystretch}{1.05}
\setlength{\aboverulesep}{0.2ex}
\setlength{\belowrulesep}{0.25ex}
\begin{tabular*}{\columnwidth}{@{\extracolsep{\fill}}lccccc@{}}
\toprule
Dataset
& \texttt{[M]}
& \texttt{[V]}
& \shortstack{Noise on\\\texttt{[M]}}
& $\Delta P_{\mathrm{GT}}$
& $\Delta$Acc. \\
\midrule
\multirow{4}{*}{Liver CT}
& CE & NC & 0\% & -0.55 & -26.47 \\
\cmidrule{2-6}
& NC & CE & 0\% & \textbf{-4.51} & \textbf{-42.65} \\
& NC & CE & 25\% & -3.23 & -35.29 \\
& NC & CE & 50\% & -1.97 & -23.53 \\
\midrule
\multirow{4}{*}{Brain MRI}
& CE & NC & 0\% & -1.00 & -13.96 \\
\cmidrule{2-6}
& NC & CE & 0\% & \textbf{-2.69} & \textbf{-19.29} \\
& NC & CE & 25\% & -2.20 & -19.80 \\
& NC & CE & 50\% & -1.76 & -21.07 \\
\midrule
\multirow{4}{*}{Breast MRI}
& CE & NC & 0\% & -1.10 & -13.33 \\
\cmidrule{2-6}
& NC & CE & 0\% & \textbf{-3.43} & \textbf{-29.63} \\
& NC & CE & 25\% & -2.66 & -25.93 \\
& NC & CE & 50\% & -1.87 & -23.70 \\
\bottomrule
\end{tabular*}
\end{table}

\noindent\textbf{Analysis protocol.}
We select paired questions for which the frozen SFT model produces an incorrect answer from the NC image and a correct answer from the CE image. The two branches independently generate their evidence and answers. At each decoder layer, we average attention over heads and next-token prediction positions within each branch. We measure the fraction of attention assigned to image tokens and the cosine similarity between the resulting spatial attention maps at corresponding foreground positions. Foreground masks exclude black background regions, and comparisons are restricted to positions valid in both images.

\noindent\textbf{Results across datasets.}
We observe similar patterns on the brain and breast MRI datasets (Figs.~\ref{fig:attention_brats} and~\ref{fig:attention_mamamia}). Despite differences in answer correctness, the NC and CE branches exhibit closely matched layer-wise trends in the fraction of attention assigned to image tokens. Mean foreground cosine similarities are 0.715 and 0.711 on brain MRI for Qwen3-VL-4B and MedGemma-4B, respectively, and 0.694 and 0.735 on breast MRI. Together with the liver CT results in the main text, these findings indicate that similar response-averaged spatial attention can coexist with different answer outcomes across anatomical sites and imaging modalities, even when the two branches generate their reasoning trajectories independently. This consistency suggests that similarity in where models attend does not necessarily imply comparable answer correctness, motivating further examination of the visual information retrieved at those positions.

\subsection{Extended Analysis of Attention and Value Interventions}
\label{app:paired_interventions}

\noindent\textbf{Subset selection and reverse-subset analysis.}
Our intervention analysis aims to identify which component more strongly mediates the model's answer advantage under one image condition over the other. We therefore focus on NC-incorrect/CE-correct pairs, where the model answers incorrectly from the NC view but correctly from the paired CE view. This correctness contrast allows us to examine whether exchanging spatial attention weights or visual values more effectively recovers the NC answer and disrupts the CE answer. Pairs answered correctly or incorrectly under both conditions remain valuable for studying prediction changes, but do not provide the same contrast for examining the transfer of a correctness advantage.

Although CE images generally improve overall answer accuracy, this aggregate advantage does not imply that every individual question benefits from the CE view. We therefore complement the primary analysis with NC-correct/CE-incorrect pairs across all three datasets using Qwen3-VL-4B, examining whether the stronger effects of visual-value exchange persist when the model's correctness advantage is reversed. As shown in Table~\ref{tab:qwen_reverse_exchange}, visual-value exchanges again produce larger changes in reference-answer probability and accuracy than attention-weight exchanges. Table~\ref{tab:qwen_reverse_noise} further shows that visual-value exchange with 25\% or 50\% noise applied to the recipient's attention weights produces larger changes in reference-answer probability than attention-weight exchange without noise, with accuracy changes generally following the same pattern. This comparison further supports the stronger influence of visual values, even when spatial attention is substantially perturbed. Together with the primary analysis, these results reveal a consistent intervention pattern across all three datasets and both directions of the correctness contrast: visual-value exchanges exert a stronger influence on answer probability and accuracy than attention-weight exchanges. This consistency supports the generality of the finding across the evaluated conditions, irrespective of whether the correct prediction comes from the NC or CE branch.

\noindent\textbf{Interventions across datasets.}
We extend the main-text intervention analysis to brain and breast MRI using Qwen3-VL-4B and MedGemma-4B. Following the same protocol, interventions are applied across decoder layers and evaluated on NC-incorrect/CE-correct subsets. Tables~\ref{tab:paired_exchange_brats} and~\ref{tab:paired_exchange_mamamia} show that exchanging visual values \texttt{[V]} produces larger changes in reference-answer probability and accuracy than exchanging spatial attention weights \texttt{[M]} in both transfer directions. Tables~\ref{tab:noisy_attention_brats} and~\ref{tab:noisy_attention_mamamia} further show that CE value transfer continues to improve NC predictions under 25\% and 50\% attention noise, although accuracy gains decrease as noise increases. These results corroborate the main-text findings across datasets and model architectures.

\begin{table}[t]
\centering
\caption{\textbf{Paired \texttt{[M]} and \texttt{[V]} exchanges on brain MRI.} Changes are reported in percentage points. Arrows indicate donor-to-recipient transfer; red $\Delta$ entries report \texttt{[V]} minus \texttt{[M]}.}
\label{tab:paired_exchange_brats}
\small
\setlength{\tabcolsep}{2pt}
\renewcommand{\arraystretch}{0.95}
\setlength{\aboverulesep}{0.2ex}
\setlength{\belowrulesep}{0.25ex}
\begin{tabular*}{\linewidth}{@{\extracolsep{\fill}}lccccc@{}}
\toprule
\multirow{2}{*}{Model}
& \multirow{2}{*}{Swap}
& \multicolumn{2}{c}{$\Delta P_{\mathrm{GT}}$}
& \multicolumn{2}{c}{$\Delta$Acc.} \\
\cmidrule(lr){3-4}\cmidrule(lr){5-6}
& & CE$\to$NC & NC$\to$CE & CE$\to$NC & NC$\to$CE \\
\midrule
\multirow{3}{*}{Qwen3-VL}
& \texttt{[M]} & +3.85 & -1.51 & +15.78 & -6.92 \\
& \texttt{[V]}
& \textbf{+8.42} & \textbf{-3.05}
& \textbf{+26.57} & \textbf{-10.79} \\
& $\Delta$
& \textcolor{gainred}{+4.57} & \textcolor{gainred}{-1.54}
& \textcolor{gainred}{+10.79} & \textcolor{gainred}{-3.86} \\
\midrule
\multirow{3}{*}{MedGemma}
& \texttt{[M]} & +7.18 & -3.35 & +22.26 & -11.50 \\
& \texttt{[V]}
& \textbf{+12.06} & \textbf{-6.18}
& \textbf{+46.17} & \textbf{-22.45} \\
& $\Delta$
& \textcolor{gainred}{+4.88} & \textcolor{gainred}{-2.83}
& \textcolor{gainred}{+23.91} & \textcolor{gainred}{-10.95} \\
\bottomrule
\end{tabular*}
\end{table}

\begin{table}[t]
\centering
\caption{\textbf{Value transfer under attention perturbations on brain MRI.} Sources of \texttt{[M]} and \texttt{[V]} are listed in the first two columns. Noise affects only \texttt{[M]} in the NC branch. Changes from original NC predictions are reported in percentage points.}
\label{tab:noisy_attention_brats}
\small
\setlength{\tabcolsep}{2pt}
\renewcommand{\arraystretch}{0.95}
\setlength{\aboverulesep}{0.2ex}
\setlength{\belowrulesep}{0.25ex}
\begin{tabularx}{\linewidth}{
    @{}ccc
    *{4}{>{\centering\arraybackslash}X}
    @{}
}
\toprule
\multirow{2}{*}{\texttt{[M]}}
& \multirow{2}{*}{\texttt{[V]}}
& \multirow{2}{*}{\shortstack{Noise on\\\texttt{[M]}}}
& \multicolumn{2}{c}{\textit{Qwen3-VL}}
& \multicolumn{2}{c}{\textit{MedGemma}} \\
\cmidrule(lr){4-5}\cmidrule(lr){6-7}
& & & $\Delta P_{\mathrm{GT}}$ & $\Delta$Acc.
    & $\Delta P_{\mathrm{GT}}$ & $\Delta$Acc. \\
\midrule
CE & NC & 0\%
& +3.85 & +15.78 & +7.18 & +22.26 \\
\midrule
NC & CE & 0\%
& \textbf{+8.42} & \textbf{+26.57}
& \textbf{+12.06} & \textbf{+46.17} \\
NC & CE & 25\%
& +7.84 & +24.80 & +12.38 & +45.26 \\
NC & CE & 50\%
& +7.46 & +24.64 & +12.68 & +43.25 \\
\bottomrule
\end{tabularx}
\end{table}

\begin{table}[t]
\centering
\caption{\textbf{Paired \texttt{[M]} and \texttt{[V]} exchanges on breast MRI.} Changes are reported in percentage points. Arrows indicate donor-to-recipient transfer; red $\Delta$ entries report \texttt{[V]} minus \texttt{[M]}.}
\label{tab:paired_exchange_mamamia}
\small
\setlength{\tabcolsep}{2pt}
\renewcommand{\arraystretch}{0.95}
\setlength{\aboverulesep}{0.2ex}
\setlength{\belowrulesep}{0.25ex}
\begin{tabular*}{\linewidth}{@{\extracolsep{\fill}}lccccc@{}}
\toprule
\multirow{2}{*}{Model}
& \multirow{2}{*}{Swap}
& \multicolumn{2}{c}{$\Delta P_{\mathrm{GT}}$}
& \multicolumn{2}{c}{$\Delta$Acc.} \\
\cmidrule(lr){3-4}\cmidrule(lr){5-6}
& & CE$\to$NC & NC$\to$CE & CE$\to$NC & NC$\to$CE \\
\midrule
\multirow{3}{*}{Qwen3-VL}
& \texttt{[M]} & +3.31 & -2.89 & +14.68 & -13.21 \\
& \texttt{[V]}
& \textbf{+7.57} & \textbf{-6.81}
& \textbf{+41.65} & \textbf{-24.95} \\
& $\Delta$
& \textcolor{gainred}{+4.27} & \textcolor{gainred}{-3.92}
& \textcolor{gainred}{+26.97} & \textcolor{gainred}{-11.74} \\
\midrule
\multirow{3}{*}{MedGemma}
& \texttt{[M]} & +6.22 & -3.74 & +21.22 & -9.23 \\
& \texttt{[V]}
& \textbf{+8.99} & \textbf{-7.23}
& \textbf{+54.61} & \textbf{-32.47} \\
& $\Delta$
& \textcolor{gainred}{+2.77} & \textcolor{gainred}{-3.49}
& \textcolor{gainred}{+33.39} & \textcolor{gainred}{-23.25} \\
\bottomrule
\end{tabular*}
\end{table}

\begin{table}[t]
\centering
\caption{\textbf{Value transfer under attention perturbations on breast MRI.} Sources of \texttt{[M]} and \texttt{[V]} are listed in the first two columns. Noise affects only \texttt{[M]} in the NC branch. Changes from original NC predictions are reported in percentage points.}
\label{tab:noisy_attention_mamamia}
\small
\setlength{\tabcolsep}{2pt}
\renewcommand{\arraystretch}{0.95}
\setlength{\aboverulesep}{0.2ex}
\setlength{\belowrulesep}{0.25ex}
\begin{tabularx}{\linewidth}{
    @{}ccc
    *{4}{>{\centering\arraybackslash}X}
    @{}
}
\toprule
\multirow{2}{*}{\texttt{[M]}}
& \multirow{2}{*}{\texttt{[V]}}
& \multirow{2}{*}{\shortstack{Noise on\\\texttt{[M]}}}
& \multicolumn{2}{c}{\textit{Qwen3-VL}}
& \multicolumn{2}{c}{\textit{MedGemma}} \\
\cmidrule(lr){4-5}\cmidrule(lr){6-7}
& & & $\Delta P_{\mathrm{GT}}$ & $\Delta$Acc.
    & $\Delta P_{\mathrm{GT}}$ & $\Delta$Acc. \\
\midrule
CE & NC & 0\%
& +3.31 & +14.68 & +6.22 & +21.22 \\
\midrule
NC & CE & 0\%
& \textbf{+7.57} & \textbf{+41.65}
& \textbf{+8.99} & \textbf{+54.61} \\
NC & CE & 25\%
& +7.58 & +36.15 & +8.95 & +51.29 \\
NC & CE & 50\%
& +7.41 & +32.84 & +8.99 & +44.28 \\
\bottomrule
\end{tabularx}
\vspace{-4mm}
\end{table}

\begin{figure*}[!htbp]
\centering
\includegraphics[width=\textwidth]{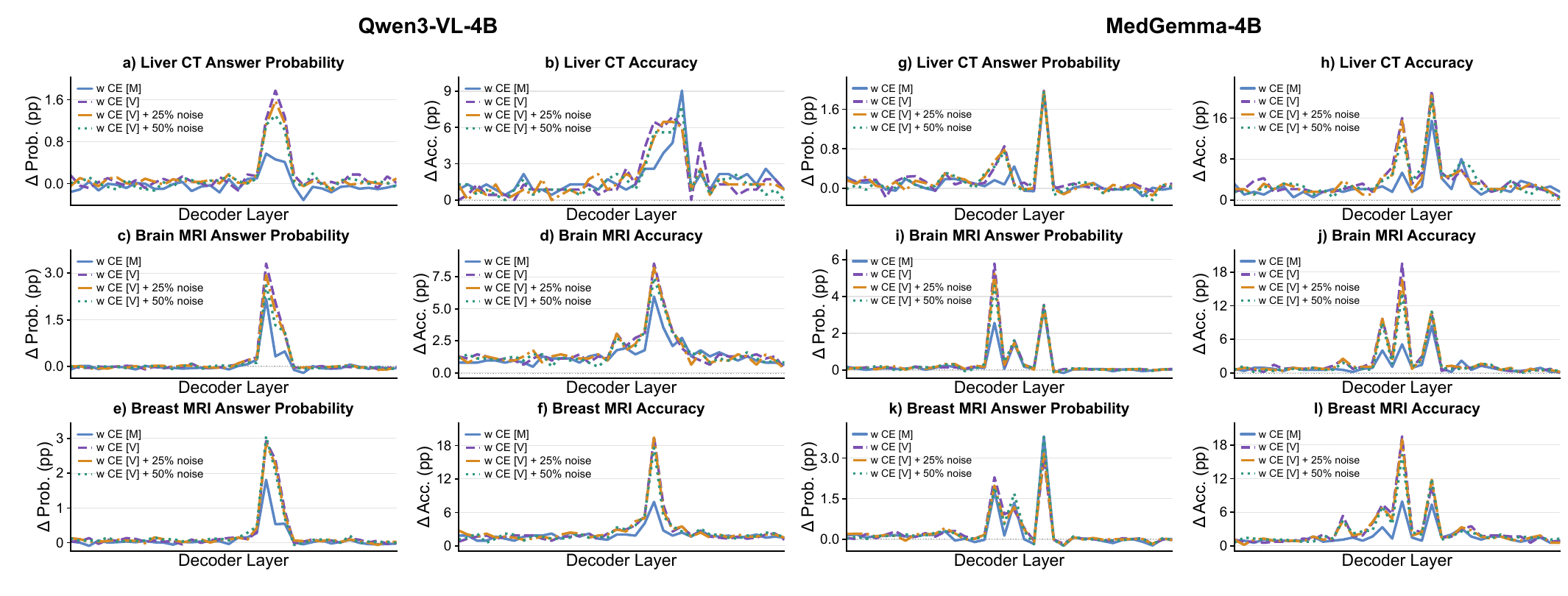}
\caption{\textbf{Layer-wise effects of attention and value interventions.} On NC-incorrect/CE-correct pairs, CE spatial attention weights \texttt{[M]} or visual values \texttt{[V]} are transferred to the NC branch at one decoder layer at a time. Curves report changes in reference-answer probability and accuracy relative to the original NC predictions, in percentage points. Noise is applied only to attention weights during value transfer.}
\label{fig:layerwise_interventions}
\end{figure*}

\noindent\textbf{Layer-wise intervention effects.}
Figure~\ref{fig:layerwise_interventions} examines CE-to-NC transfer on the NC-incorrect/CE-correct subsets by intervening at one decoder layer at a time. Across datasets and models, visual-value exchanges produce more pronounced response peaks than attention-weight exchanges in both reference-answer probability and accuracy. Adding noise to attention weights generally weakens these gains while preserving a substantial advantage over attention-weight transfer around the dominant peaks. The single-layer results thus reinforce the all-layer finding that visual values exert a stronger influence on answer correctness than spatial attention weights. The effects are also strongly depth-dependent: the largest changes concentrate in the middle layers, whereas early and final layers exhibit comparatively small responses. This concentration is consistent with prior findings on the role of middle layers in visual--language fusion~\cite{yin2025clearsight}, suggesting that transferring enhanced visual values is particularly effective at stages where visual information is integrated into task-relevant representations.

\subsection{Consistency across Training Regimes}
\label{app:consistency}

To examine the consistency of our findings across training regimes, we independently fine-tune two Qwen3-VL-4B models, one on NC images and the other on CE images, and repeat the paired analysis on liver CT. This setting allows each model to adapt separately to its respective image condition. During independently generated responses, the two models exhibit a mean attention map cosine similarity of 0.651, showing that spatial attention similarity persists despite separate fine-tuning. Table~\ref{tab:separate_exchange} further shows that visual-value exchanges \texttt{[V]} produce larger bidirectional changes in reference-answer probability and accuracy than spatial attention-weight exchanges \texttt{[M]}.  Table~\ref{tab:separate_noise} shows that this pattern remains evident under attention perturbations. Even with 50\% noise applied to the recipient's attention weights, CE value transfer improves NC reference-answer probability and accuracy by 3.72 and 28.28 percentage points, respectively, exceeding the gains from attention-weight exchange without noise. Together, these results reproduce the main intervention pattern in separately trained models and support its consistency across training regimes.

\begin{table}[!htbp]
\centering
\caption{\textbf{Paired \texttt{[M]} and \texttt{[V]} exchanges between separately trained models on liver CT.}
Both branches are Qwen3-VL-4B models independently fine-tuned on NC or CE images.
Results are reported in percentage points.
Arrows indicate donor-to-recipient transfer.
Red $\Delta$ entries report \texttt{[V]} minus \texttt{[M]}.}
\label{tab:separate_exchange}
\small
\setlength{\tabcolsep}{3pt}
\renewcommand{\arraystretch}{0.95}
\setlength{\aboverulesep}{0.2ex}
\setlength{\belowrulesep}{0.25ex}
\resizebox{0.98\linewidth}{!}{%
\begin{tabular}{@{\hspace{3pt}}lccccc@{\hspace{3pt}}}
\toprule
\multirow{2}{*}{Model}
& \multirow{2}{*}{Swap}
& \multicolumn{2}{c}{$\Delta P_{\mathrm{GT}}$}
& \multicolumn{2}{c}{$\Delta$Acc.} \\
\cmidrule(lr){3-4}\cmidrule(lr){5-6}
& & CE$\to$NC & NC$\to$CE & CE$\to$NC & NC$\to$CE \\
\midrule
\multirow{3}{*}{Qwen3-VL}
& \texttt{[M]} & +1.07 & -1.25 & +21.21 & -10.77 \\
& \texttt{[V]}
& \textbf{+4.53} & \textbf{-3.30}
& \textbf{+33.00} & \textbf{-19.87} \\
& $\Delta$
& \textcolor{gainred}{+3.45}
& \textcolor{gainred}{-2.05}
& \textcolor{gainred}{+11.78}
& \textcolor{gainred}{-9.09} \\
\bottomrule
\end{tabular}%
}
\end{table}

\begin{table}[!htbp]
\centering
\caption{\textbf{Value transfer under attention perturbations.}
Results use separately trained Qwen3-VL-4B models on liver CT.
Changes relative to the NC baseline are reported in percentage points.
Noise is applied to NC attention during CE value transfer.}
\label{tab:separate_noise}
\small
\setlength{\tabcolsep}{3pt}
\renewcommand{\arraystretch}{1.08}
\setlength{\aboverulesep}{0.2ex}
\setlength{\belowrulesep}{0.25ex}

\begin{tabularx}{\linewidth}{
    @{}l*{4}{>{\centering\arraybackslash}X}@{}
}
\toprule
\multirow{2}{*}{Metric}
& CE \texttt{[M]}
& \multicolumn{3}{c}{CE \texttt{[V]}} \\
\cmidrule(lr){2-2}\cmidrule(lr){3-5}
& No noise & No noise & 25\% noise & 50\% noise \\
\midrule
$\Delta P_{\mathrm{GT}}$
& +1.07 & \textbf{+4.53} & +4.13 & +3.72 \\
$\Delta$Acc.
& +21.21 & \textbf{+33.00} & +30.98 & +28.28 \\
\bottomrule
\end{tabularx}
\end{table}

\section{Additional Experimental Results}
\label{app:experiments}

\subsection{Training configuration}
\label{app:implementation}

Table~\ref{tab:training_config} summarizes the shared training configurations used by our method and the compared on-policy distillation baselines. We first fine-tune the instruction-tuned models for two epochs using combined NC and CE data. All on-policy distillation methods are initialized from the same SFT checkpoints and trained for one additional epoch. Both stages use $512\times512$ images and LoRA applied to decoder linear layers, with rank 64, scaling factor 128, and dropout 0.05. All methods use AdamW with a learning rate of $5\times10^{-5}$, with a cosine schedule during SFT and a constant learning rate during post-training.

\begin{table}[t]
\centering
\caption{\textbf{Training configurations for MLLMs.} SFT uses combined NC and CE data; On-policy Distillation methods use NC rollouts with a frozen CE teacher.}
\label{tab:training_config}
\small
\setlength{\tabcolsep}{3pt}
\renewcommand{\arraystretch}{0.95}
\setlength{\aboverulesep}{0.2ex}
\setlength{\belowrulesep}{0.25ex}
\begin{tabularx}{\columnwidth}{@{}l*{2}{>{\centering\arraybackslash}X}@{}}
\toprule
Configuration & SFT & On-policy Distillation \\
\midrule
\rowcolor{blue!4}\multicolumn{3}{c}{\textit{Image Input}} \\
Image size & $512\times512$ & $512\times512$ \\
\midrule
\rowcolor{blue!4}\multicolumn{3}{c}{\textit{LoRA}} \\
Rank / alpha & 64 / 128 & 64 / 128 \\
Dropout & 0.05 & 0.05 \\
Target modules & Decoder linear & Decoder linear \\
\midrule
\rowcolor{blue!4}\multicolumn{3}{c}{\textit{Optimization}} \\
Initialization & Instruct & 2-epoch SFT \\
Training epochs & 2 & 1 \\
Optimizer & AdamW & AdamW \\
Learning rate & $5\times10^{-5}$ & $5\times10^{-5}$ \\
LR schedule & Cosine & Constant \\
\bottomrule
\end{tabularx}
\end{table}

\subsection{Robustness Across Random Seeds and Question Types}
\label{app:breakdown}

\noindent\textbf{Stability across random seeds.}
Table~\ref{tab:seed_variability} reports mean accuracy and standard deviation across three training seeds on the liver CT dataset using Qwen3-VL-2B. Our method outperforms all compared baselines on both the internal and external test sets under every evaluated seed, achieving mean accuracies of $66.54\pm0.35\%$ and $52.17\pm0.47\%$, respectively. These results demonstrate a consistent performance advantage across seeds, with small standard deviations indicating limited sensitivity to the training seed under the evaluated setting.

\noindent\textbf{Consistency across question types and option counts.}
Aggregate accuracy may obscure differences across question categories, making it important to examine whether performance gains are concentrated in particular tasks. For instance, in the liver CT dataset, lesion identification, lesion characterization, and multi-region localization questions have two, three, and four answer options, respectively. This correspondence allows the same breakdown to characterize performance jointly across question types and their associated answer formats. Table~\ref{tab:liver_option_count} reports these results at both model scales. Our method achieves the highest or joint-highest accuracy in ten of the twelve comparisons and ranks second in the remaining two, including the highest or joint-highest accuracy in all six external-test comparisons. Although the magnitude of improvement varies across tasks and model scales, the consistently strong external-test results indicate that the overall advantage spans all three question categories. This finer-grained evaluation complements aggregate accuracy by showing how performance varies across distinct visual assessment tasks.

\begin{table}[t]
\centering
\caption{\textbf{Across-seed performance on the liver CT dataset with Qwen3-VL-2B.} We report mean accuracy $\pm$ standard deviation on the full internal and external test sets. Best and second-best results are \textbf{bold} and \underline{underlined}, respectively.}
\label{tab:seed_variability}
\small
\setlength{\tabcolsep}{2pt}
\renewcommand{\arraystretch}{1.15}
\setlength{\aboverulesep}{0.4ex}
\setlength{\belowrulesep}{0.65ex}
\resizebox{\columnwidth}{!}{%
\begin{tabular}{@{}lcccc@{}}
\toprule
Test set & OPSD & Med-OPD & VA-OPD & VIVID \\
\midrule
Internal & 62.59$\pm$1.59 & \underline{64.56$\pm$0.28} & 62.80$\pm$0.46 & \textbf{66.54$\pm$0.35} \\
External & 46.62$\pm$1.06 & \underline{47.13$\pm$0.59} & 46.23$\pm$0.48 & \textbf{52.17$\pm$0.47} \\
\bottomrule
\end{tabular}%
}
\end{table}

\begin{table}[t]
\centering
\caption{\textbf{Liver CT dataset results by option count and question category.} Accuracy (\%) on internal and external test sets. Best and second-best results are bold and underlined, respectively.}
\label{tab:liver_option_count}
\small
\setlength{\tabcolsep}{3pt}
\renewcommand{\arraystretch}{1.02}
\setlength{\aboverulesep}{0.2ex}
\setlength{\belowrulesep}{0.25ex}
\begin{tabular*}{\columnwidth}{@{\extracolsep{\fill}}lcccc@{}}
\toprule
\multirow{2}{*}{Method} & \multicolumn{2}{c}{\textit{Qwen3-VL-2B}} & \multicolumn{2}{c}{\textit{Qwen3-VL-4B}} \\
\cmidrule(lr){2-3}\cmidrule(lr){4-5}
& Internal & External & Internal & External \\
\midrule
\rowcolor{blue!4}\multicolumn{5}{c}{\textit{Lesion Identification (2 options)}} \\
OPSD & \underline{72.30} & \underline{54.32} & 77.50 & \underline{64.61} \\
Med-OPD & 72.04 & 52.47 & 78.67 & 62.96 \\
VA-OPD & 71.26 & 52.88 & \underline{78.80} & 64.20 \\
VIVID & \textbf{74.90} & \textbf{61.93} & \textbf{79.06} & \textbf{66.87} \\
\midrule
\rowcolor{blue!4}\multicolumn{5}{c}{\textit{Lesion Characterization (3 options)}} \\
OPSD & 52.68 & \textbf{38.07} & 59.73 & 40.61 \\
Med-OPD & \textbf{55.03} & \underline{37.06} & 57.72 & \underline{43.15} \\
VA-OPD & 49.33 & \underline{37.06} & \underline{60.40} & \underline{43.15} \\
VIVID & \underline{53.69} & \textbf{38.07} & \textbf{60.74} & \textbf{43.65} \\
\midrule
\rowcolor{blue!4}\multicolumn{5}{c}{\textit{Multi-region Localization (4 options)}} \\
OPSD & 52.31 & 39.41 & 59.62 & 45.88 \\
Med-OPD & 52.31 & \underline{40.59} & \textbf{61.92} & \underline{48.24} \\
VA-OPD & \underline{53.46} & 40.00 & 60.00 & 46.47 \\
VIVID & \textbf{56.15} & \textbf{42.94} & \underline{60.38} & \textbf{48.82} \\
\bottomrule
\end{tabular*}
\end{table}

\subsection{Additional Ablations}
\label{app:moreablation}

\noindent\textbf{Training efficiency.}
Table~\ref{tab:efficiency} reports computational measurements on NVIDIA H20 GPUs. Our method records lower mean training step times than Med-OPD and VA-OPD at both model scales. Med-OPD and VA-OPD estimate token-level visual dependence by comparing factual and counterfactual predictions, requiring two forward passes and thus introducing additional training computation. Our method has higher peak Actor allocated memory, reflecting the storage required for paired visual values and auxiliary-response construction.

\noindent\textbf{Inference efficiency.}
We measure inference latency at batch size one. The observed latencies are similar across methods, ranging from 1.44 to 1.53 seconds per question for 2B models and from 1.71 to 1.86 seconds for 4B models. Allocated inference memory is identical across methods within each model scale: 4.07 GiB for 2B and 8.44 GiB for 4B. Our auxiliary branch and projection-based target construction are used only during training, leaving a standard NC student at deployment.

\begin{table}[t]
\centering
\caption{\textbf{Computational efficiency using NVIDIA H20 GPUs.} Training reports historical mean step time and peak Actor allocated memory; batch configurations and retained response counts differ across runs. Inference reports mean latency at batch size one. Memory excludes separately hosted teacher processes.}
\label{tab:efficiency}
\small
\setlength{\tabcolsep}{3pt}
\renewcommand{\arraystretch}{0.95}
\setlength{\aboverulesep}{0.2ex}
\setlength{\belowrulesep}{0.25ex}
\begin{tabular*}{\columnwidth}{@{\extracolsep{\fill}}llcccc@{}}
\toprule
\multirow{2}{*}{Scale} & \multirow{2}{*}{Method} & \multicolumn{2}{c}{Training} & \multicolumn{2}{c}{Inference} \\
\cmidrule(lr){3-4}\cmidrule(lr){5-6}
& & s/step & GiB & s/question & GiB \\
\midrule
\multirow{3}{*}{2B} & Med-OPD & 86.10 & 19.88 & 1.44 & 4.07 \\
& VA-OPD & 73.49 & 16.35 & 1.53 & 4.07 \\
& VIVID & 47.46 & 30.31 & 1.53 & 4.07 \\
\midrule
\multirow{3}{*}{4B} & Med-OPD & 85.38 & 44.94 & 1.71 & 8.44 \\
& VA-OPD & 91.94 & 35.30 & 1.86 & 8.44 \\
& VIVID & 82.63 & 56.65 & 1.82 & 8.44 \\
\bottomrule
\end{tabular*}
\vspace{-4mm}
\end{table}

\noindent\textbf{Hyperparameter and objective sensitivity.}
Table~\ref{tab:sensitivity} examines foreground attention coverage $\tau$, CE value mixing ratio $\alpha$, the weak teacher-supervision coefficient $\lambda$, the distillation objective, and vocabulary support size $K$ on the liver CT dataset. Each variant changes one setting while keeping the remaining configuration fixed. We vary $\tau$ and $\alpha$ from their default of 0.5 to 0.25 or 0.75, increase $\lambda$ from 0.1 to 0.2 or 0.3, replace JSD with forward or reverse KL, and reduce $K$ from 100 to 10 or 50 while retaining the tail bucket. Reducing $\tau$ to 0.25 consistently lowers accuracy across both model scales and test sets, suggesting that restricting value mixing to a smaller set of visual positions exclude useful lesion-related information. Performance remains relatively stable under the other tested variations.


\begin{table}[t]
\centering
\caption{\textbf{Hyperparameter and objective sensitivity on liver CT.} Each variant changes one setting from the default VIVID configuration. $\tau$ denotes foreground attention coverage, $\alpha$ the CE value mixing ratio, $\lambda$ the weak teacher-supervision coefficient, and $K$ the vocabulary support size excluding the retained tail bucket. Accuracy is reported in percentages, with the best result in each column shown in \textbf{bold}.}
\label{tab:sensitivity}
\small
\setlength{\tabcolsep}{3pt}
\renewcommand{\arraystretch}{0.95}
\setlength{\aboverulesep}{0.2ex}
\setlength{\belowrulesep}{0.25ex}
\begin{tabular*}{\columnwidth}{@{\extracolsep{\fill}}lcccc@{}}
\toprule
\multirow{2}{*}{Variant} & \multicolumn{2}{c}{\textit{Qwen3-VL-2B}} & \multicolumn{2}{c}{\textit{Qwen3-VL-4B}} \\
\cmidrule(lr){2-3}\cmidrule(lr){4-5}
& Internal & External & Internal & External \\
\midrule
\rowcolor{blue!4}\multicolumn{5}{c}{\textit{Attention Coverage}} \\
$\tau=0.25$ & 65.56 & 50.88 & 70.99 & 56.39 \\
$\tau=0.75$ & 66.16 & 51.82 & 70.91 & 57.44 \\
\midrule
\rowcolor{blue!4}\multicolumn{5}{c}{\textit{Value Mixing Ratio}} \\
$\alpha=0.25$ & 66.31 & 52.52 & 71.14 & 56.62 \\
$\alpha=0.75$ & 66.01 & 52.17 & \textbf{71.89} & 57.33 \\
\midrule
\rowcolor{blue!4}\multicolumn{5}{c}{\textit{Teacher-supervision Coefficient}} \\
$\lambda=0.2$ & 66.39 & 52.05 & 70.84 & 57.33 \\
$\lambda=0.3$ & 66.47 & 52.40 & 71.14 & 55.80 \\
\midrule
\rowcolor{blue!4}\multicolumn{5}{c}{\textit{Loss Function}} \\
Forward KL & \textbf{66.69} & 51.93 & 70.76 & 57.80 \\
Reverse KL & 66.39 & 50.41 & 70.31 & 56.15 \\
\midrule
\rowcolor{blue!4}\multicolumn{5}{c}{\textit{Vocabulary Support}} \\
$K=10$ & 65.18 & 50.76 & 71.14 & 57.56 \\
$K=50$ & 66.54 & 51.58 & 71.21 & 57.44 \\
\midrule
Ours & 66.47 & \textbf{52.64} & 71.29 & \textbf{57.91} \\
\bottomrule
\end{tabular*}
\end{table}

\subsection{Patch Selection and Projection Analysis}
\label{app:intervention_behavior}

We examine attention-guided patch selection and response-guided projection in the trained models using the training-time intervention configuration with $\tau=0.5$. We report the fraction of non-background patches selected for value mixing, the directional alignment between teacher corrections and auxiliary responses, and the correction magnitude retained after projection.

\noindent\textbf{Attention-guided patch selection.}
As shown in Table~\ref{tab:actual_selection_ratio}, the overall selected fraction ranges from 9.78\% to 16.12\% of non-background patches. Thus, a relatively small subset of visual positions accounts for half of the foreground attention mass. Across all datasets and model scales, the middle and late layer groups select fewer patches than the early group under the same coverage target, indicating more concentrated foreground attention in these layers. These results show that value mixing is restricted to a compact, attention-selected subset of visual positions.

\noindent\textbf{Directional alignment and retained corrections.}
Table~\ref{tab:projection_behavior} shows that the auxiliary-response direction aligns positively with the teacher correction at 79.98\%--85.02\% of generation positions. CE value mixing therefore induces a prediction shift with a positive component along the teacher correction at most positions. However, mean cosine similarities range from 0.197 to 0.247, indicating only partial directional alignment. The positive projection retains an average of 22.23\%--26.81\% of the full teacher-correction norm, including positions where the projection is zero. The resulting correction is therefore nonzero at most positions while retaining only the component aligned with the student's response to CE values.

\begin{table}[t]
\centering
\caption{\textbf{Attention-guided patch selection.} Values report selected patches as a percentage of non-background patches. Early, Middle, and Late summarize the corresponding layer groups; Overall aggregates across all layers.}
\label{tab:actual_selection_ratio}
\small
\setlength{\tabcolsep}{3pt}
\renewcommand{\arraystretch}{0.92}
\setlength{\aboverulesep}{0.2ex}
\setlength{\belowrulesep}{0.25ex}
\begin{tabular*}{\columnwidth}{
  @{\extracolsep{\fill}}
  lccc>{\columncolor{gray!8}[\tabcolsep][0pt]}c@{}
}
\toprule
Dataset & Early & Middle & Late & \makebox[0.13\columnwidth][c]{Overall} \\
\midrule
\rowcolor{blue!4}\multicolumn{5}{c}{\textit{Qwen3-VL-2B}} \\
Liver CT   & 14.68 & 9.68  & 9.10  & 11.08 \\
Brain MRI  & 18.37 & 14.92 & 15.07 & 16.08 \\
Breast MRI & 10.84 & 10.04 & 8.58  & 9.78  \\
\midrule
\rowcolor{blue!4}\multicolumn{5}{c}{\textit{Qwen3-VL-4B}} \\
Liver CT   & 14.82 & 8.49  & 6.63  & 9.98  \\
Brain MRI  & 17.79 & 15.57 & 15.00 & 16.12 \\
Breast MRI & 14.22 & 9.93  & 7.40  & 10.52 \\
\bottomrule
\end{tabular*}
\vspace{-6mm}
\end{table}

\begin{table}[!htbp]
\centering
\caption{\textbf{Response-guided projection statistics.} Cosine reports mean cosine similarity between the teacher-correction and auxiliary-response directions. Positive reports the percentage of generation positions with a nonzero positive projection. Retained norm reports the mean ratio $\|d\|_2/\|r\|_2$ as a percentage, where $r$ is the full teacher correction and $d$ its positive projection, including positions with zero projection.}
\label{tab:projection_behavior}
\small
\setlength{\tabcolsep}{3pt}
\renewcommand{\arraystretch}{0.92}
\setlength{\aboverulesep}{0.2ex}
\setlength{\belowrulesep}{0.25ex}
\begin{tabularx}{\columnwidth}{
  @{}l*{3}{>{\centering\arraybackslash}X}@{}
}
\toprule
Dataset & Cosine & Positive (\%) & \shortstack{Retained (\%)} \\
\midrule
\rowcolor{blue!4}\multicolumn{4}{c}{\textit{Qwen3-VL-2B}} \\
Liver CT   & 0.224 & 83.61 & 24.54 \\
Brain MRI  & 0.208 & 82.13 & 23.53 \\
Breast MRI & 0.247 & 85.02 & 26.81 \\
\midrule
\rowcolor{blue!4}\multicolumn{4}{c}{\textit{Qwen3-VL-4B}} \\
Liver CT   & 0.197 & 80.43 & 22.23 \\
Brain MRI  & 0.229 & 84.89 & 24.99 \\
Breast MRI & 0.208 & 79.98 & 23.81 \\
\bottomrule
\end{tabularx}
\end{table}

\begin{figure*}[t]
\centering
\includegraphics[width=\textwidth]{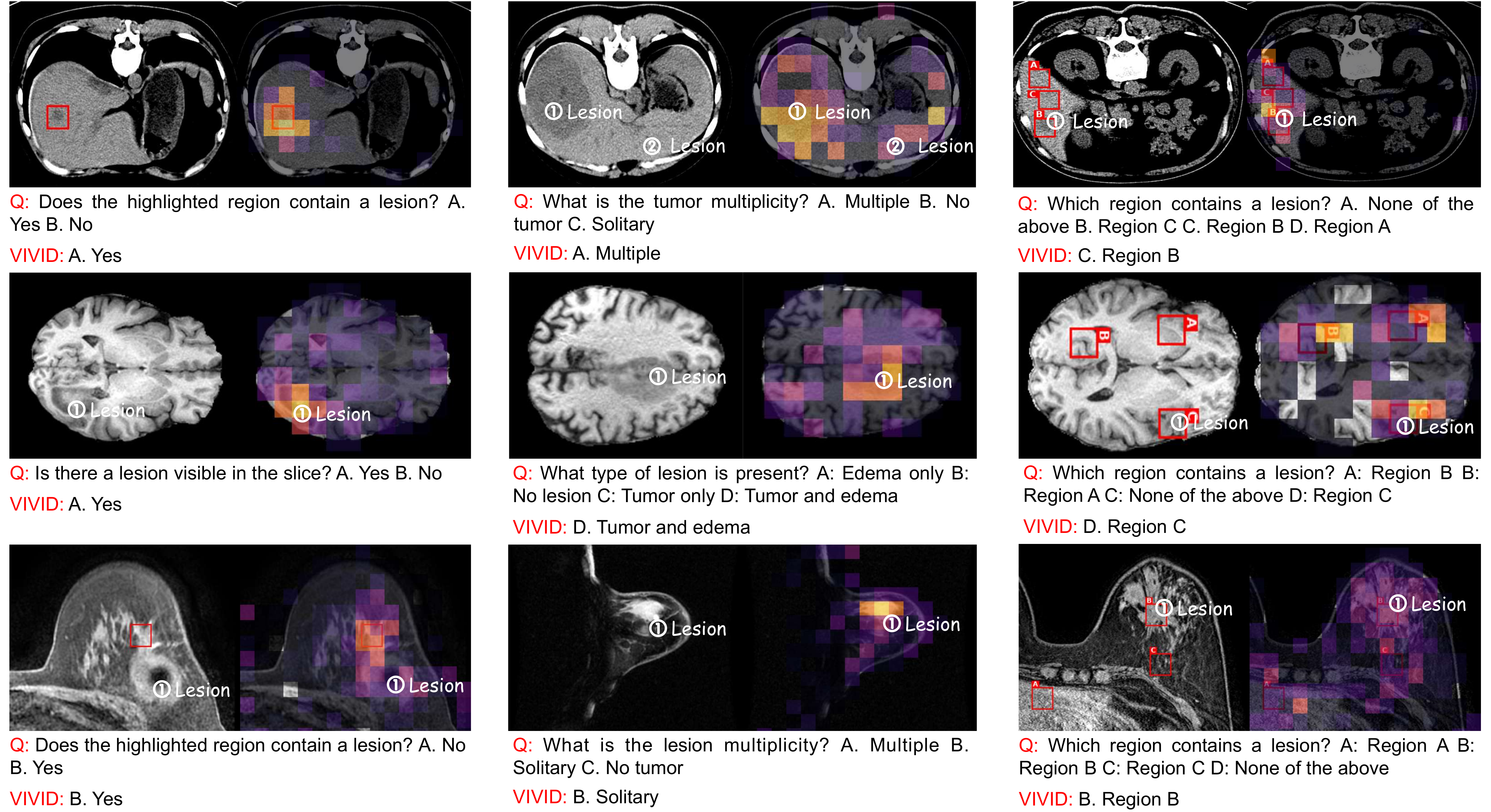}
\caption{\textbf{Spatial distribution of patches selected for value mixing.} Rows show liver CT, brain MRI, and breast MRI examples; columns illustrate lesion identification, lesion characterization, and local visual discrimination. Each example presents the NC image (left) and its patch-selection heatmap (right), followed by the question and VIVID's predicted answer. Heatmaps aggregate selection frequencies across decoder layers and generation positions under $\tau=0.5$; brighter colors indicate more frequent selection. Red boxes denote queried or candidate regions, and numbered lesion labels provide spatial references.}
\label{fig:patch_selection_maps}
\end{figure*}

\noindent\textbf{Spatial distribution of selected patches.}
Fig.~\ref{fig:patch_selection_maps} visualizes patch selection across the three datasets. Each heatmap reports the selection frequency of non-background patches across decoder layers and generation positions, with brighter colors indicating more frequent selection for CE value mixing. In the illustrated cases, high-frequency selections are predominantly concentrated within or near the task-relevant organ. For questions involving highlighted or candidate regions, frequently selected patches tend to cluster around these regions. For slice-level lesion identification, prominent selections occur near the marked lesions. These qualitative spatial correspondences suggest that attention-guided selection directs value mixing toward anatomically and question-relevant regions.

\begin{table}[!htbp]
\centering
\caption{\textbf{Five-point rubric for generated evidence correctness.} Ratings assess factual agreement with the reference evidence, including lesion presence, attributes, location, and candidate-region relationships. Semantically equivalent wording is accepted.}
\label{tab:generated_evidence_rubric}
\small
\setlength{\tabcolsep}{6pt}
\renewcommand{\arraystretch}{1.12}
\begin{tabularx}{\columnwidth}{@{}c >{\raggedright\arraybackslash}X@{}}
\toprule
\textbf{Score} & \textbf{Assessment criteria} \\
\midrule
5 & Core facts agree with the reference, with no substantive contradictions. Paraphrases and omissions that do not affect the factual meaning are acceptable. \\
\addlinespace[5pt]
4 & Core facts agree, but minor details are omitted, descriptions are less precise, or additional details cannot be verified from the reference. \\
\addlinespace[5pt]
3 & The evidence partially agrees with the reference but omits key information, is too vague to establish the main finding, or introduces substantial unsupported details. \\
\addlinespace[5pt]
2 & The evidence substantively contradicts the reference regarding lesion presence, multiplicity, composition, location, candidate regions, or important imaging characteristics. \\
\addlinespace[5pt]
1 & The core facts are predominantly opposite to the reference, the content is irrelevant, or no usable evidence is provided. \\
\bottomrule
\end{tabularx}
\end{table}

\begin{table}[!htbp]
\centering
\caption{\textbf{Five-point rubric for evidence--answer consistency.} Ratings assess whether the generated evidence supports the model's own predicted answer, independently of reference correctness. Reference answers and evidence are not shown during this assessment.}
\label{tab:evidence_answer_rubric}
\small
\setlength{\tabcolsep}{6pt}
\renewcommand{\arraystretch}{1.12}
\begin{tabularx}{\columnwidth}{@{}c >{\raggedright\arraybackslash}X@{}}
\toprule
\textbf{Score} & \textbf{Assessment criteria} \\
\midrule
5 & The evidence clearly and sufficiently supports the predicted answer, with no internal contradictions. \\
\addlinespace[5pt]
4 & The evidence generally supports the predicted answer, but contains a minor reasoning gap or an imprecise statement of scope. \\
\addlinespace[5pt]
3 & The evidence is compatible with the predicted answer but is too vague or incomplete to distinguish it from alternative options. \\
\addlinespace[5pt]
2 & The evidence mainly supports another answer or contains a substantial contradiction, although some support for the predicted answer remains. \\
\addlinespace[5pt]
1 & The evidence directly contradicts the predicted answer, or no usable evidence is provided. \\
\bottomrule
\end{tabularx}
\end{table}

\subsection{Generated Evidence Quality and Case Studies}
\label{app:expert_evaluation}

\begin{table}[!htbp]
\centering
\caption{\textbf{Correctness of model-generated evidence.} Scores are reported as mean \textpm{} standard deviation. Higher is better; the highest mean score for each dataset is shown in \textbf{bold}.}
\label{tab:evidence_correctness}
\small
\setlength{\tabcolsep}{3pt}
\renewcommand{\arraystretch}{1.15}
\resizebox{\columnwidth}{!}{%
\begin{tabular}{@{}lccccc@{}}
\toprule
Dataset & SFT & OPSD & Med-OPD & VA-OPD & VIVID \\
\midrule
Liver
& 4.01\textpm1.10
& 4.03\textpm1.10
& 4.06\textpm1.00
& 4.05\textpm1.02
& \textbf{4.17\textpm0.89} \\
Brain
& 4.72\textpm0.64
& 4.65\textpm0.61
& 4.68\textpm0.60
& 4.68\textpm0.60
& \textbf{4.74\textpm0.58} \\
Breast
& 4.16\textpm0.93
& 4.14\textpm0.94
& 4.14\textpm0.90
& 4.10\textpm0.95
& \textbf{4.18\textpm0.94} \\
\bottomrule
\end{tabular}%
}
\end{table}

\noindent\textbf{Evaluation protocol.}
For each dataset, a radiologist evaluates 200 randomly sampled questions for each of two complementary assessments: evidence correctness and evidence--answer consistency. For evidence correctness, we sample from the common subset of questions answered correctly by all compared methods, allowing evidence quality to be compared on the same questions while controlling for answer correctness. Generated evidence is assessed against the reference evidence, allowing semantically equivalent descriptions rather than requiring exact wording. For evidence--answer consistency, sampling is not restricted to correctly answered questions. The physician considers only the question, answer options, generated evidence, and the model's predicted answer, without access to the reference answer or evidence. An incorrect answer can therefore receive a high consistency score if it is clearly supported by the generated evidence. Both assessments use five-point scales, detailed in Tables~\ref{tab:generated_evidence_rubric} and~\ref{tab:evidence_answer_rubric}.

\noindent\textbf{Evaluation results.}
Evidence--answer consistency scores are almost uniformly at the maximum across methods, indicating that the generated evidence generally supports each model's predicted answer. Table~\ref{tab:evidence_correctness} further reports evidence correctness on the shared correctly answered subsets. VIVID obtains the highest mean score on all three benchmarks, with the largest gain on liver CT, suggesting favorable evidence quality alongside its answer accuracy.

\begin{figure*}[t]
\centering
\includegraphics[width=\textwidth]{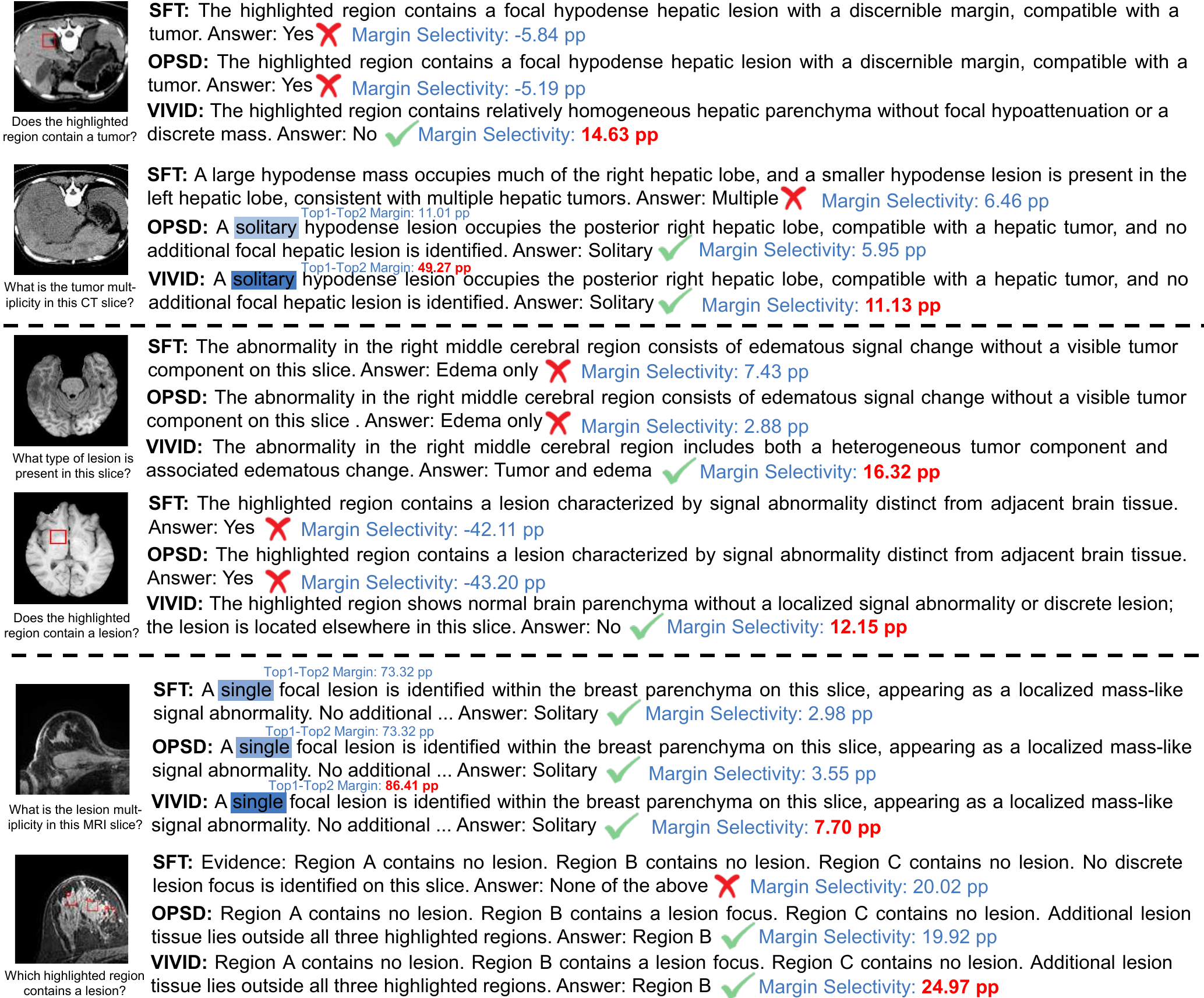}
\caption{\textbf{Qualitative comparisons of SFT, OPSD, and VIVID across three datasets.} Examples show model-generated evidence and answers for liver CT, brain MRI, and breast MRI. Checkmarks and crosses indicate correct and incorrect answers, respectively. Margin selectivity is reported alongside each response, and top-1--top-2 probability margins are shown for highlighted evidence tokens, both in percentage points. VIVID corrects baseline errors and exhibits higher margin selectivity in these examples. Even when methods produce the same correct answer, VIVID assigns a larger probability margin to the highlighted answer-relevant token.}
\label{fig:case_studies}
\end{figure*}

\noindent\textbf{Qualitative case studies.}
Fig.~\ref{fig:case_studies} compares representative outputs from SFT, OPSD, and VIVID across the three benchmarks. VIVID corrects baseline errors in lesion presence, multiplicity, and composition, producing evidence that more accurately describes the queried findings. In the highlighted-region examples, VIVID correctly identifies lesion-negative queried regions on lesion-positive slices and keeps its evidence aligned with the spatial scope of the question. It also achieves higher margin selectivity ($m_S$) in all displayed cases. Even when methods produce the same correct answer, VIVID exhibits a larger top-1--top-2 probability margin at the highlighted first task-relevant token, such as ``single'' or ``solitary.'' This indicates a stronger local preference for the correct lesion description, revealing differences in predictive certainty that are not captured by answer accuracy alone.


{
    \small
    \bibliographystyle{ieeenat_fullname}
    \bibliography{main}
}

\end{document}